\documentclass{article}

\usepackage{subcaption}
\usepackage{tcolorbox}
\usepackage{float}

\usepackage{graphicx}
\usepackage{pifont}
\usepackage{booktabs}
\usepackage{multirow}
\usepackage[table]{xcolor}
\usepackage{makecell}
\usepackage{booktabs}
\usepackage{colortbl}
\usepackage{amsmath}
\usepackage{enumitem}

\definecolor{ForestGreen}{RGB}{34,139,34}
\definecolor{BrickRed}{RGB}{178,34,34}

\definecolor{mygreen}{rgb}{0.0, 0.7, 0.0}
\newcommand{\cmark}{\textcolor{mygreen}{\ding{51}}}
\newcommand{\xmark}{\textcolor{red}{\ding{55}}}
\PassOptionsToPackage{numbers, compress}{natbib}

\definecolor{grayhead}{HTML}{F2F2F2}
\definecolor{polishc}{HTML}{E8F1FF}
\definecolor{parac}{HTML}{EEF7E8}
\definecolor{stylec}{HTML}{FFF4D6}
\definecolor{compressc}{HTML}{FDE8E8}
\definecolor{expandc}{HTML}{EDE7F6}

 \usepackage[preprint]{neurips_2026}

\usepackage[utf8]{inputenc} 
\usepackage[T1]{fontenc}    
\usepackage{hyperref}       
\usepackage{url}           
\usepackage{booktabs}      
\usepackage{amsfonts}       
\usepackage{nicefrac}       
\usepackage{microtype}     
\usepackage{xcolor}        

\title{Operation-Guided Progressive Human-to-AI Text Transformation Benchmark for Multi-Granularity AI-Text Detection}

\author{
\begin{tabular}{c}
Sondos Mahmoud Bsharat$^{1}$ \quad
Jiacheng Liu$^{1}$ \quad
Xiaohan Zhao$^{1}$ \quad
Tianjun Yao$^{1}$ \\
Xinyi Shang$^{1,2}$ \quad
Yi Tang$^{1}$ \quad 
Jiacheng Cui$^{1}$ \quad
Ahmed Elhagry$^{1}$  \\
Salwa K. Al Khatib$^{1}$ \quad
Hao Li$^{1}$ \quad
Salman Khan$^{1}$ \quad
Zhiqiang Shen$^{1,\dagger}$\\
\\[-0.5em]
$^{1}$ Mohamed bin Zayed University of Artificial Intelligence \\
$^{2}$ University College London\\
$^{\dagger}$Correspondence: \texttt{zhiqiang.shen@mbzuai.ac.ae}
\end{tabular}
}

\begin{document}

\maketitle

\begin{abstract}
As AI writing assistants become increasingly integrated into real-world drafting and revision workflows, many documents are no longer purely human-written or AI-generated, but instead result from progressive human–AI co-editing. However, existing AI-text detection benchmarks largely focus on final outputs and provide limited understanding of how AI authorship signals emerge, accumulate, or disappear throughout the revision process. We introduce \textbf{OpAI-Bench}, an operation-guided benchmark for studying progressive human-to-AI text transformation across document, sentence, token, and span granularities. Starting from human-written documents, OpAI-Bench constructs nine sequentially revised versions for each sample under predefined AI coverage levels and five representative AI edit operations, covering four domains while preserving complete authorship provenance at multiple granularities. The benchmark supports comprehensive evaluation with 8 document-level detectors, 7 sentence-level detectors, and 2 fine-grained token/span-level detectors. Experiments reveal that AI-text detectability is governed not only by the proportion of AI-edited content, but also by edit operation, domain, and cumulative revision history. Interestingly, we notice that mixed-authorship intermediate versions are often harder to detect than both fully human and heavily AI-edited endpoints, exposing non-monotonic detection patterns missed by existing benchmarks. OpAI-Bench provides a controlled testbed for analyzing \textit{whether}, \textit{when}, and \textit{how} AI-assisted writing becomes detectable under realistic progressive editing scenarios. Our code and benchmark are available at \url{https://github.com/VILA-Lab/OpAI-Bench}.

\end{abstract}

\section{Introduction}
AI-assisted writing and editing workflows~\citep{saha-feizi-2025-almost,thai2026editlens,uchendu-etal-2021-turingbench-benchmark, guo2023close,he2024mgtbench,macko-etal-2023-multitude} are increasingly embedded in practical writing workflows, where they are used not only to generate complete passages, but also to revise, polish, expand, compress, and restructure human-written drafts. As a result, many real-world documents are better characterized as products of progressive human–AI co-editing rather than as purely human-written or fully AI-generated text. This emerging writing paradigm challenges the conventional binary framing of AI-text detection, where a document is assumed to belong to one of two endpoint categories. In practice, AI involvement may appear locally, accumulate gradually, and interact with prior human content across multiple rounds of revision.

Existing AI-text detection benchmarks~\cite{dugan-etal-2024-raid,wang-etal-2024-m4, wang-etal-2024-m4gt} provide valuable resources for evaluating whether a completed text is human- or AI-written, but they offer limited support for analyzing how detectability evolves during the transformation from human writing to AI-edited text. Most benchmarks are constructed from static final outputs and do not preserve the intermediate revision states that lead to those outputs. Consequently, they cannot answer several important questions: at what stage does AI involvement become reliably detectable, which types of edits introduce the strongest detection signals, and whether mixed-authorship texts are easier or harder to detect than endpoint cases. These questions are increasingly important as AI-assisted writing becomes more incremental, interactive, and operation-specific.

Evaluation should capture not only how much text is AI-edited, but also how it is edited and where the edits occur. Edit operations can leave different signals: polishing may preserve lexical choices while improving fluency, paraphrasing changes form while retaining meaning, expansion adds explanatory detail, and compression removes details or simplifies structure. Collapsing these operations into a single AI-written label obscures their different effects on detection. Similarly, document-level labels support global screening but cannot localize AI involvement, while sentence-level labels may miss mixed human--AI fragments within a sentence. Token- and span-level provenance is therefore needed to evaluate fine-grained localization.

To address these limitations, we introduce {\bf OpAI-Bench}, a novel operation-guided benchmark for progressive human-to-AI text transformation and multi-granularity AI-text detection. Starting from human-written source documents, OpAI-Bench constructs a sequence of progressively revised versions under predefined AI coverage levels and representative edit operations. Each revision trajectory records how human text is transformed by AI edits over time, while preserving authorship provenance at the document, sentence, token, and span levels. This design enables controlled evaluation of detection performance as a function of AI coverage, edit operation, domain, and cumulative revision history, rather than only on isolated final texts.

Using OpAI-Bench, we conduct a comprehensive evaluation of document-level, sentence-level, and fine-grained AI-text detectors. Our results show that AI-text detectability is not determined solely by the proportion of AI-edited content. Instead, detector performance varies substantially across edit operations, domains, and revision stages. In particular, intermediate mixed-authorship versions can be more challenging than both purely human and heavily AI-edited endpoints, revealing non-monotonic detection behavior that is overlooked by existing static benchmarks. These findings suggest that reliable AI-text detection requires moving beyond binary endpoint classification toward trajectory-aware and operation-aware evaluation.

Our contributions are summarized as follows:
\begin{itemize}
    \item We introduce {\bf OpAI-Bench}, an operation-guided benchmark for progressive human-to-AI text transformation that preserves intermediate revision states rather than only final outputs.

    \item We construct cumulative revision trajectories with predefined AI coverage levels and five edit operations: \textit{polish}, \textit{paraphrase}, \textit{style rewrite}, \textit{compress}, and \textit{expand}, and provide provenance across document, sentence, token, and span levels.

    \item We benchmark diverse detector families across granularities and use OpAI-Bench to assess detector stability across coverage, edit operations, domains, generators, and revision history. Our results reveal non-monotonic detectability, with a critical mixed-authorship region around $v_4$ where intermediate AI coverage and compression coincide.
\end{itemize}

\section{Related Work}

\noindent{\bf AI-text detection benchmarks.}
Prior work on AI-text detection has introduced increasingly diverse benchmarks spanning human-written, machine-generated, and AI-edited text. Benchmarks such as TuringBench~\citep{uchendu-etal-2021-turingbench-benchmark}, HC3~\citep{guo2023close}, MGTBench~\citep{he2024mgtbench}, MULTITuDE~\cite{macko-etal-2023-multitude}, RAID~\citep{dugan-etal-2024-raid}, M4~\citep{wang-etal-2024-m4}, and M4GT-Bench~\citep{wang-etal-2024-m4gt} broaden evaluation across generators, domains, languages, and robustness conditions. More recent benchmarks also expand the problem formulation to include robustness under attacks, generator attribution, and mixed-authorship boundary detection. DetectRL~\citep{NEURIPS2024_b61bdf7e} evaluates detectors under prompt attacks, paraphrasing, perturbations, and data-mixing settings, while M4GT-Bench~\citep{wang-etal-2024-m4gt} further incorporates generator attribution and mixed human--machine change-point detection. However, these benchmarks still mainly evaluate the final text obtained after generation or editing, or at most a single authorship transition within it, rather than a full revision process with explicit intermediate stages.

\noindent{\bf Mixed-authorship and fine-grained attribution.}
A related line of work moves beyond pure human-versus-AI classification to study hybrid and coauthored text. MixSet~\citep{zhang-etal-2024-llm} highlights the difficulty of detecting mixed human--AI writing in settings such as AI-revised human drafts and human-revised machine outputs. Localization methods such as AdaLoc~\citep{zhang-etal-2024-machine} identify machine-generated sentences within otherwise human-written documents, while SenDetEX~\citep{jiang-etal-2025-sendetex}, HACo-Det~\citep{su-etal-2025-haco}, and DAMASHA~\citep{teja-etal-2026-damasha} move toward finer attribution at the sentence, word, or token levels. Beemo~\citep{artemova-etal-2025-beemo} and RealBench~\citep{he2025detree} further show that human post-editing and diverse collaboration patterns substantially complicate detection. These works demonstrate the importance of mixed-authorship and fine-grained detection, but typically focus on a completed hybrid sample or local attribution within it, rather than the revision trajectory through which mixed authorship is formed.

\noindent{\bf Varying degrees of AI involvement.}
Recent work has begun to move beyond binary AI-text detection by considering text with varying degrees of AI involvement. APT-Eval~\citep{saha-feizi-2025-almost} varies the degree of AI polishing and shows that lightly edited text can be difficult to distinguish from fully human writing. PaLD~\citep{lei2025pald} estimates the proportion and localization of LLM-written content, while EditLens~\citep{thai2026editlens} models AI involvement as a continuous edit-magnitude signal relative to an original human draft. HACo-Det~\citep{su-etal-2025-haco} similarly motivates numeric notions of AI ratio through fine-grained attribution. Together, these works move beyond binary labels, but they typically focus on final edited samples or single editing outcomes. In contrast, OpAI-Bench studies explicit cumulative revision trajectories with preserved intermediate versions, controlled AI coverage, and diverse edit operations.

\begin{figure*}[t]
  \centering
    \includegraphics[width=\textwidth]{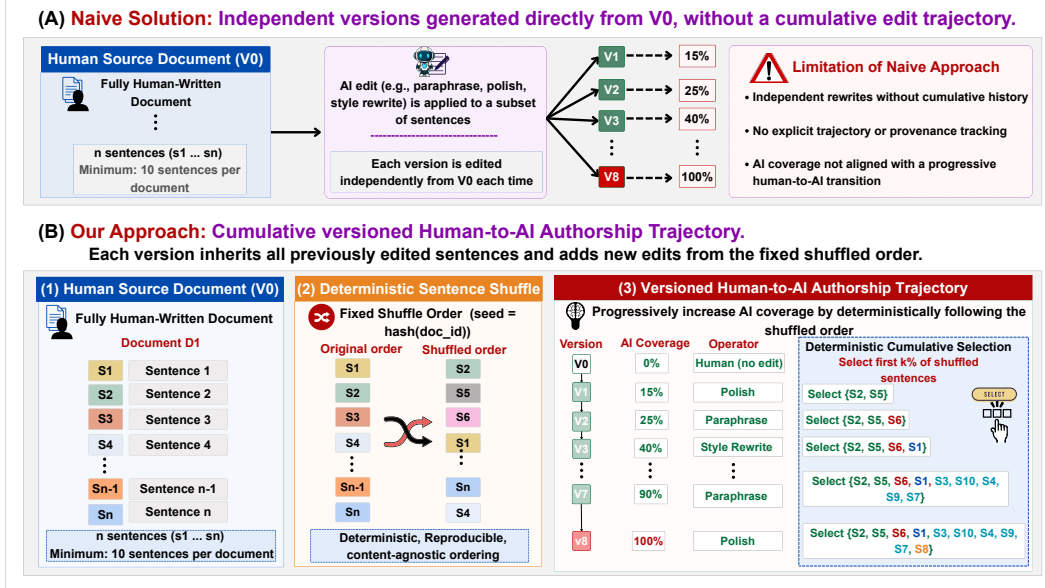}
    \vspace{-0.1in}
\caption{OpAI-Bench construction pipeline. Top: a naive setup creates each version independently from the original human document, so versions do not form a true revision history and provenance is not tracked across stages. Bottom: OpAI-Bench builds a cumulative trajectory by progressively editing a larger prefix of that order. Each version is generated from the previous one.}
  \label{ben}
  \vspace{-0.15in}
\end{figure*}

\vspace{-0.1in}
\section{OpAI-Bench}
\label{benchmark_construction}
\vspace{-0.05in}
\subsection{Overview and Design Principles}
\label{overview}

Most prior AI-text detection benchmarks evaluate a completed text and assign it a final authorship label, such as human, AI-generated, or mixed. This formulation has been valuable for static detection, but it is less informative for studying how detectability evolves as AI involvement is introduced through revision. Prior benchmarks generally do not preserve explicit intermediate versions, do not model cumulative version-to-version editing, and do not jointly control AI coverage, edit type, and multi-granularity provenance within a unified setup. Consequently, they provide only limited support for analyzing when detection becomes reliable, whether matched AI coverage yields different difficulty across edit types, and whether detector behavior depends on revision history as well as on the final edited text.

OpAI-Bench is designed to study this setting by representing mixed authorship as a controlled revision trajectory. Given a source document $D$, we construct
\(
\mathcal{T}(D) = \bigl(D^{(0)}, D^{(1)}, \dots, D^{(8)}\bigr),
\)
where $D^{(0)}$ is the original human-written document and $D^{(V)}$ denotes version $V \in \{0,\dots,8\}$. Rather than generating each version independently from the original source, we construct each $D^{(V)}$ by editing the previous version $D^{(V-1)}$. This yields a cumulative sequence in which the target AI coverage increases monotonically from fully human (\texttt{v0}) to fully AI-edited (\texttt{v8}) under predefined coverage ratios and edit operations.

Each version is paired with provenance annotations at the token, sentence, and document levels. This structure enables analyses that are difficult to carry out on static benchmarks, including coverage-controlled comparisons, edit-type-controlled comparisons, and trajectory-controlled comparisons. In particular, it allows us to study whether detectability depends only on the final amount of AI-edited content, or also on the path through which mixed authorship is formed. Table~\ref{tab:benchmark-comparison} situates OpAI-Bench relative to prior resources, and Figure~\ref{ben} illustrates the construction pipeline.

\begin{table}[t]
\centering
\small
\setlength{\tabcolsep}{4.3pt}
\begin{tabular}{lcc|cc|ccc|c}
\toprule
\textbf{Benchmark} &
\multicolumn{2}{c|}{\textbf{Authorship}} &
\multicolumn{2}{c|}{\textbf{Trajectory}} &
\multicolumn{3}{c|}{\textbf{Diversity}} &
\textbf{Annotation} \\
\cmidrule(lr){2-3} \cmidrule(lr){4-5} \cmidrule(lr){6-8} \cmidrule(lr){9-9}
&
\textbf{Mixed} &
\textbf{AI\_Cov} &
\textbf{Interm} &
\textbf{Cumul} &
\textbf{Edit\_Types} &
\textbf{Domain} &
\textbf{LLM} &
\textbf{Gran.} \\
\midrule
DetectRL \cite{NEURIPS2024_b61bdf7e} & \cmark & \xmark & \xmark & \xmark & \cmark & \cmark & \cmark & D \\
MixSet \citep{zhang-etal-2024-llm} & \cmark & \xmark & \xmark & \xmark & \cmark & \cmark & \cmark & M \\
RAID \citep{dugan-etal-2024-raid} & \xmark & \xmark & \xmark & \xmark & \cmark & \cmark & \cmark  & D \\
M4GT-Bench \citep{wang-etal-2024-m4gt} & \cmark & \xmark & \xmark & \xmark & \xmark & \cmark & \cmark  & M \\
RealBench \citep{he2025detree} & \cmark & \xmark & \xmark & \xmark & \cmark & \cmark & \cmark  & D \\
Beemo \citep{artemova-etal-2025-beemo} & \cmark & \xmark & \xmark & \xmark & \cmark & \cmark & \cmark & D \\
HACo-Det \citep{su-etal-2025-haco} & \cmark & \xmark & \xmark & \cmark & \xmark & \cmark & \cmark  &  M\\
SenDetEX \citep{jiang-etal-2025-sendetex} & \cmark & \xmark & \xmark & \xmark & \xmark &\cmark& \cmark & S \\
APT-Eval \citep{saha-feizi-2025-almost} & \cmark & \cmark & \xmark & \xmark & \xmark & \cmark & \cmark & D  \\
DAMASHA \citep{teja-etal-2026-damasha} & \cmark & \xmark & \xmark & \xmark & \xmark & \cmark & \cmark & B \\
EditLens \citep{thai2026editlens} & \cmark & \xmark & \xmark & \xmark & \cmark & \cmark & \cmark  & D \\
\midrule
\textbf{OpAI-Bench (Ours)} & \cmark & \cmark & \cmark & \cmark & \cmark & \cmark & \cmark  & M \\
\bottomrule
\end{tabular}
 \vspace{0.1in}

\caption{Comparison with prior AI-involved text detection benchmarks. Mixed: mixed human--AI authorship; AI\_Cov: controlled AI coverage; Interm: intermediate revision stages; Cumul: cumulative version-to-version revision; Edit\_Types: diverse AI edit operations; Gran.: supported annotation/evaluation granularity (D: document, S: sentence, B: boundary, M: multi-granularity). OpAI-Bench is the only benchmark listed that combines all properties.}
\label{tab:benchmark-comparison}
\vspace{-0.25in}
\end{table}

OpAI-Bench is guided by five design principles: \textbf{versioned trajectories}, expanding each source document into nine ordered versions (\texttt{v0}--\texttt{v8}) from fully human to fully AI-edited text; \textbf{controlled AI coverage}, introducing edits at predefined sentence-level ratios from 0\% to 100\%; \textbf{edit-type diversity}, covering \textit{polish}, \textit{paraphrase}, \textit{style rewrite}, \textit{compress}, and \textit{expand}; \textbf{cumulative construction}, generating each version from the previous one rather than independently from the source; and \textbf{multi-granularity provenance}, preserving AI-revision authorship labels at word-level token/span, sentence, and document levels.

\subsection{Benchmark Construction}

\noindent{\bf Source Domains, Representation, and Filtering.} OpAI-Bench is constructed from human-written documents across four domains: student essays~\cite{learning_agency_lab_aes2}, news articles~\cite{Narayan2018DontGM}, government reports~\cite{huang-etal-2021-efficient}, and scientific abstracts~\cite{arxiv_paper_abstracts_kaggle}. These domains vary in writing style, discourse structure, and sentence length, enabling evaluation across diverse text types.

We use \emph{document} as the unit of editing. Depending on the source corpus, a document may be a multi-paragraph text or a single paragraph treated as a document. Each document is normalized and segmented into paragraphs and sentences, represented as \(D=(p_1,\dots,p_m)\), where each paragraph \(p_i=(s^{(i)}_1,\dots,s^{(i)}_{n_i})\). Stable paragraph and sentence identifiers are assigned and preserved across all derived versions for sentence selection, reconstruction, and authorship tracking.

To support intermediate revision trajectories, we retain only documents with at least 10 sentences, i.e., \(N(D)=\sum_{i=1}^{m} n_i \geq 10\). Each retained document is then expanded into a nine-version trajectory while preserving its paragraph and sentence structure. Summary statistics are reported in Table~\ref{tab:dataset_stats}.

\begin{table}[t]
\centering
\resizebox{0.94\linewidth}{!}{
\begin{tabular}{l c c c c c}
\toprule
\textbf{Domain} & \textbf{\# Source Docs} & \textbf{\# Traj.} & \textbf{\# Versioned Samples} & \textbf{Avg.\ sents} & \textbf{Avg.\ tokens} \\
\midrule
Student essays       & 3,969 & 7,906 & 71,154 & 21.0 & 398.8 \\
News articles        & 3,998 & 7,892 & 71,028 & 24.0 & 491.3 \\
Government reports   & 3,993 & 8,000 & 72,000 & 20.6 & 563.7 \\
Scientific abstracts & 3,762 & 7,291 & 65,612 & 11.0 & 234.3 \\
\midrule
\textbf{Total} & 15,722 & 31,089 & 279,794 & 19.3 & 426.1 \\
\bottomrule
\end{tabular}
}
 \vspace{0.05in}
\caption{Statistics of the OpAI-Bench main split. Source documents are distinct human-written \texttt{v0} texts, while trajectories are generator-specific revision sequences initialized from \texttt{v0}. Versioned samples count all released versions from \texttt{v0} to \texttt{v8} across the train, development, and test splits. Statistics are reported for the three primary generators and exclude ablation splits.}
\label{tab:dataset_stats}
\vspace{-0.13in}
\end{table}

\noindent{\bf Versioned Trajectories and Cumulative Editing.} For each source document \(D\), OpAI-Bench constructs an ordered trajectory
\(\mathcal{T}(D)=(D^{(0)},D^{(1)},\dots,D^{(8)})\), where \(D^{(0)}\) is the original human-written document and later versions follow the fixed operation--coverage schedule in Table~\ref{tab:schedule}. The trajectory spans the transition from fully human text at \texttt{v0} to fully AI-edited text at \texttt{v8}, with intermediate versions increasing the amount of edited content while varying the edit operation.
To select edited sentences, we define a fixed document-specific ordering $\pi(D) = (\pi_1, \ldots, \pi_{N(D)})$ using a deterministic shuffle seeded by the document identifier. This 
ordering is computed once and reused across all versions, ensuring reproducibility and removing positional bias. Given target sentence-level coverage $c^{(t)}_\text{sent}$ at 
version $t$, we select $k^{(t)} = \lceil c^{(t)}_\text{sent} \cdot N(D) \rceil$ sentences and define the edited set as $S^{(t)} = \{\pi_1, \ldots, \pi_{k^{(t)}}\}$. Because  coverage increases monotonically, the edited sets satisfy  $S^{(0)} \subseteq S^{(1)} \subseteq \cdots \subseteq S^{(8)}$. This yields a cumulative editing process: once selected, a sentence remains editable in later versions, so subsequent versions both add newly selected sentences and re-edit previously modified ones under the current operation. This 
preserves not only where AI edits occur, but also the revision history through which they accumulate.

\noindent{\bf Editing Operations and Coverage Control.}
OpAI-Bench uses five sentence-level rewrite operations to model different forms of AI-assisted revision: \textit{polish}, \textit{paraphrase}, \textit{style rewrite}, \textit{compress}, and \textit{expand}. These operations vary the form of intervention: \textit{polish} makes fluency edits, \textit{paraphrase} changes wording while preserving meaning, \textit{style rewrite} modifies tone, \textit{compress} removes non-essential phrasing, and \textit{expand} adds limited clarifying detail. All prompts require the model to rewrite only the target sentence, preserve names, numbers, entities, and factual content, avoid unsupported additions, and return a single sentence. We further enforce sentence-level consistency through automatic validation and manual verification: outputs are discarded if they split the target into multiple sentences, merge across sentence boundaries, or violate the expected format. Full prompts are provided in Appendix~\ref{app:prompts}.

For each version $D^{(t)}$, the target sentence-level AI coverage $c^{(t)}_\text{sent}$ determines the fraction of sentences selected for rewriting according to Table~\ref{tab:schedule}. A sentence is labeled AI-edited if selected for rewriting. Token- and span-level annotations are projected from AI-edited sentence regions using whitespace tokenization: words whose character offsets 
overlap an AI-marked region are labeled AI, and consecutive AI-labeled words are merged into spans. These labels are tokenizer-agnostic and can be aligned to any subword vocabulary via character offsets. This convention is 
consistent with segment-level mixed-authorship settings that label AI-revised regions as machine-involved~\citep{zhang-etal-2024-llm}.

\begin{table}[t]
\centering
\footnotesize
\setlength{\tabcolsep}{9pt}
\begin{tabular}{c*{9}{c}}
\toprule
\rowcolor{grayhead}
\textbf{Version} & $v_0$ & $v_1$ & $v_2$ & $v_3$ & $v_4$ & $v_5$ & $v_6$ & $v_7$ & $v_8$ \\
\midrule
\textbf{Operation} 
& none 
& \cellcolor{polishc} polish 
& \cellcolor{parac} para. 
& \cellcolor{stylec} style 
& \cellcolor{compressc} compress 
& \cellcolor{expandc} expand 
& \cellcolor{stylec} style 
& \cellcolor{parac} para. 
& \cellcolor{polishc} polish \\
\textbf{Coverage} 
& 0\% & 15\% & 25\% & 40\% & \cellcolor{compressc} 50\% & 60\% & 75\% & 90\% & 100\% \\
\bottomrule
\end{tabular}
\vspace{0.05in}
\caption{OpAI-Bench version schedule. Each version is defined by an edit operation and target sentence-level AI coverage. ``para.'' denotes paraphrase.}
\label{tab:schedule}
\vspace{-0.22in}
\end{table}

\vspace{-0.07in}
\section{Experiments}
\vspace{-0.07in}

\begin{table}[t]
\centering
\resizebox{0.93\linewidth}{!}{  
\begin{tabular}{llcccccccc}
\toprule
\textbf{Detector} & \textbf{Ver.} & \multicolumn{2}{c}{\textbf{Essays}} & \multicolumn{2}{c}{\textbf{Reports}} & \multicolumn{2}{c}{\textbf{News}} & \multicolumn{2}{c}{\textbf{Abstracts}} \\
\cmidrule(lr){3-4} \cmidrule(lr){5-6} \cmidrule(lr){7-8} \cmidrule(lr){9-10}
 &  & \textbf{Acc} & \textbf{F1-AI} & \textbf{Acc} & \textbf{F1-AI} & \textbf{Acc} & \textbf{F1-AI} & \textbf{Acc} & \textbf{F1-AI} \\
\midrule
 \multicolumn{10}{c}{\textbf{Sentence Level}} \\
\midrule

\multirow{9}{*}{Claude-Haiku} & $v0$ & 91.2 & 0.0 & 96.4 & 0.0 & 91.9 & 0.0 & 53.2 & 0.0 \\
 & $v1$ & \cellcolor{BrickRed!10}77.2 {\scriptsize \textcolor{BrickRed}{-13.9}} & \cellcolor{ForestGreen!15}22.0 {\scriptsize \textcolor{ForestGreen}{+22.0}} & \cellcolor{BrickRed!12}79.9 {\scriptsize \textcolor{BrickRed}{-16.5}} & \cellcolor{ForestGreen!8}7.0 {\scriptsize \textcolor{ForestGreen}{+7.0}} & \cellcolor{BrickRed!10}78.0 {\scriptsize \textcolor{BrickRed}{-13.9}} & \cellcolor{ForestGreen!8}11.2 {\scriptsize \textcolor{ForestGreen}{+11.2}} & \cellcolor{BrickRed!8}50.6 {\scriptsize \textcolor{BrickRed}{-2.6}} & \cellcolor{ForestGreen!18}25.9 {\scriptsize \textcolor{ForestGreen}{+25.9}} \\
 & $v2$ & \cellcolor{BrickRed!8}74.7 {\scriptsize \textcolor{BrickRed}{-2.5}} & \cellcolor{ForestGreen!16}45.5 {\scriptsize \textcolor{ForestGreen}{+23.5}} & \cellcolor{BrickRed!8}69.4 {\scriptsize \textcolor{BrickRed}{-10.5}} & \cellcolor{ForestGreen!8}11.4 {\scriptsize \textcolor{ForestGreen}{+4.4}} & \cellcolor{BrickRed!8}70.5 {\scriptsize \textcolor{BrickRed}{-7.5}} & \cellcolor{ForestGreen!8}20.3 {\scriptsize \textcolor{ForestGreen}{+9.1}} & \cellcolor{BrickRed!8}48.9 {\scriptsize \textcolor{BrickRed}{-1.6}} & \cellcolor{ForestGreen!8}37.6 {\scriptsize \textcolor{ForestGreen}{+11.7}} \\
\bottomrule
\end{tabular}
}
\vspace{0.08in}
\caption{Aggregate main results across revision versions and domains, averaged over generators. We report accuracy and F1-AI for zero-shot detectors and LLM-as-detectors at their native evaluation granularity: document, sentence, or fine-grained token/span level. Colored deltas show the change from the previous version.}
\label{tab:main_table_aggregate}
\vspace{-0.27in}
\end{table}

\subsection{Tasks}

OpAI-Bench evaluates AI-text detection at three granularities over the same versioned trajectory $\mathcal{T}(D)=(D^{(0)},\dots,D^{(8)})$. \textbf{Document-level detection} predicts whether a document version contains any AI-edited content. The source version $D^{(0)}$ is labeled human, while later versions are labeled AI-involved. \textbf{Sentence-level attribution} predicts a binary authorship label for each sentence, marking whether the sentence has entered the cumulative AI-edited set by the corresponding version. \textbf{Token- and span-level localization} evaluates fine-grained provenance by identifying the AI-edited portions of a document at the benchmark token or character-span level.

All tasks are evaluated at each version $t \in \{0,\dots,8\}$ rather than only at the final edited document. This allows us to measure detector behavior throughout the human-to-AI revision trajectory.

\subsection{Evaluation Protocol}

\noindent{\bf Evaluation regimes.}
We organize detectors into three categories corresponding to distinct levels of task-specific adaptation. \textbf{(i)~Zero-shot methods}: detectors are evaluated in their original settings, using either released checkpoints or models reproduced from the original training scripts, with no exposure to OpAI-Bench data. This measures how well existing detection methods transfer to trajectory-style mixed authorship without adaptation. Document-level methods include Desklib~\citep{desklib_ai_text_detector_v101}, DetectLLM~\citep{su2023detectllm}, E5-Small~\citep{wang2022text}, Fast-DetectGPT~\citep{bao2023fast}, OOD-LLM Detect~\citep{zeng2025human}, RADAR~\citep{hu2023radar}, RoBERTa OpenAI~\citep{solaiman2019release}, and GigaCheck~\citep{tolstykh2024gigacheck}; sentence-level methods include AdaLoc~\citep{zhang-etal-2024-machine}, GenAI Sentence~\citep{teja2025fine}, GL-CLiC~\citep{adi-etal-2025-gl}, and SeqXGPT~\citep{wang-etal-2023-seqxgpt}; token- and span-level methods include DAMASHA~\citep{teja-etal-2026-damasha} and GigaCheck~\citep{tolstykh2024gigacheck}.

\textbf{(ii)~LLM-as-detector}: three frontier language models: GPT-5.4, Gemini~3~Flash, and Claude~Haiku~4.5 are queried as zero-shot classifiers via sentence-level confidence prompting, without any gradient-based adaptation; they serve as prompt-based reference points.

\textbf{(iii)~Trained on OpAI-Bench}: detectors are fine-tuned on the OpAI-Bench training split using sentence-level provenance labels drawn from three in-distribution generators (GPT-5.4, GPT-5.4-nano, Gemini~2.5~Flash). Qwen3-8B is withheld from training and evaluated as a held-out generator, isolating cross-generator transfer while holding the training protocol fixed. Document-level methods include Fast-DetectGPT and GigaCheck; sentence-level methods include AdaLoc, GenAI-Sentence, GL-CLiC, and SeqXGPT; token-level and span-level methods include DAMASHA and GigaCheck. Full implementation details and training hyperparameters are provided in the appendix. 

\noindent{\bf Domain and generator splits.}
We report results by domain and generator to evaluate robustness across writing settings and generation backends. For fine-tuned methods, Qwen3-8B is used only as a held-out generator for cross-generator evaluation.

\noindent{\bf Metrics.}
For each task and version, we report accuracy and AI-class $F_1$ ($F_1^{\mathrm{AI}}$). Main tables also show changes relative to the previous reported version, and the ablation studies examine coverage control, edit operations, and cumulative construction.

\begin{figure*}[t]
    \centering
    \includegraphics[width=0.99\textwidth]
    {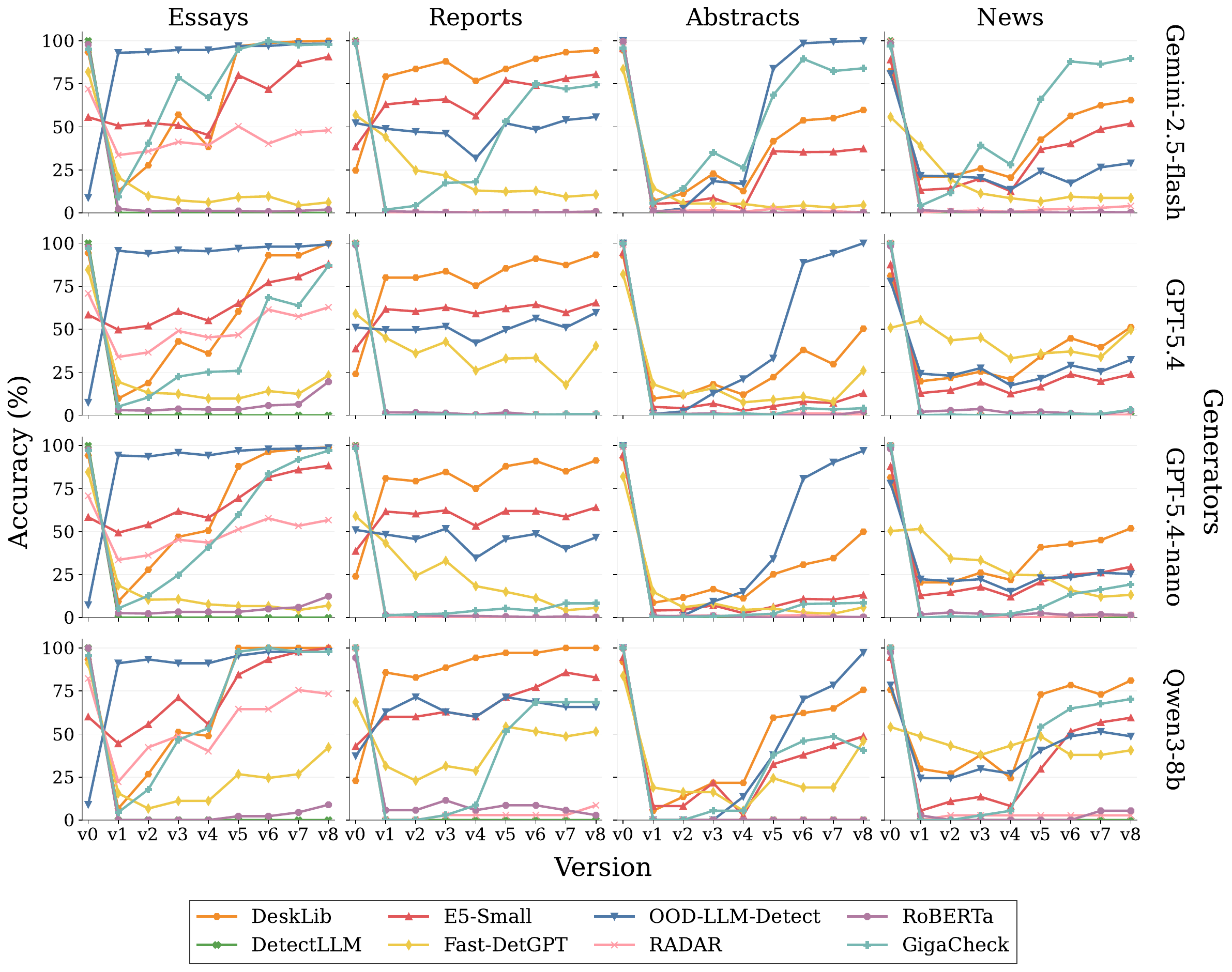}
\caption{Document-level accuracy across revision versions, domains, and generators. Each curve represents a document-level detector, illustrating how performance changes along the progressive human-to-AI revision trajectory.}
    \label{fig:app_doc_acc_1_main}
    \vspace{-0.15in}
\end{figure*}

\section{Main Results}
\label{sec:main_results}

\begin{figure*}[t]
    \centering
    \includegraphics[width=\textwidth]{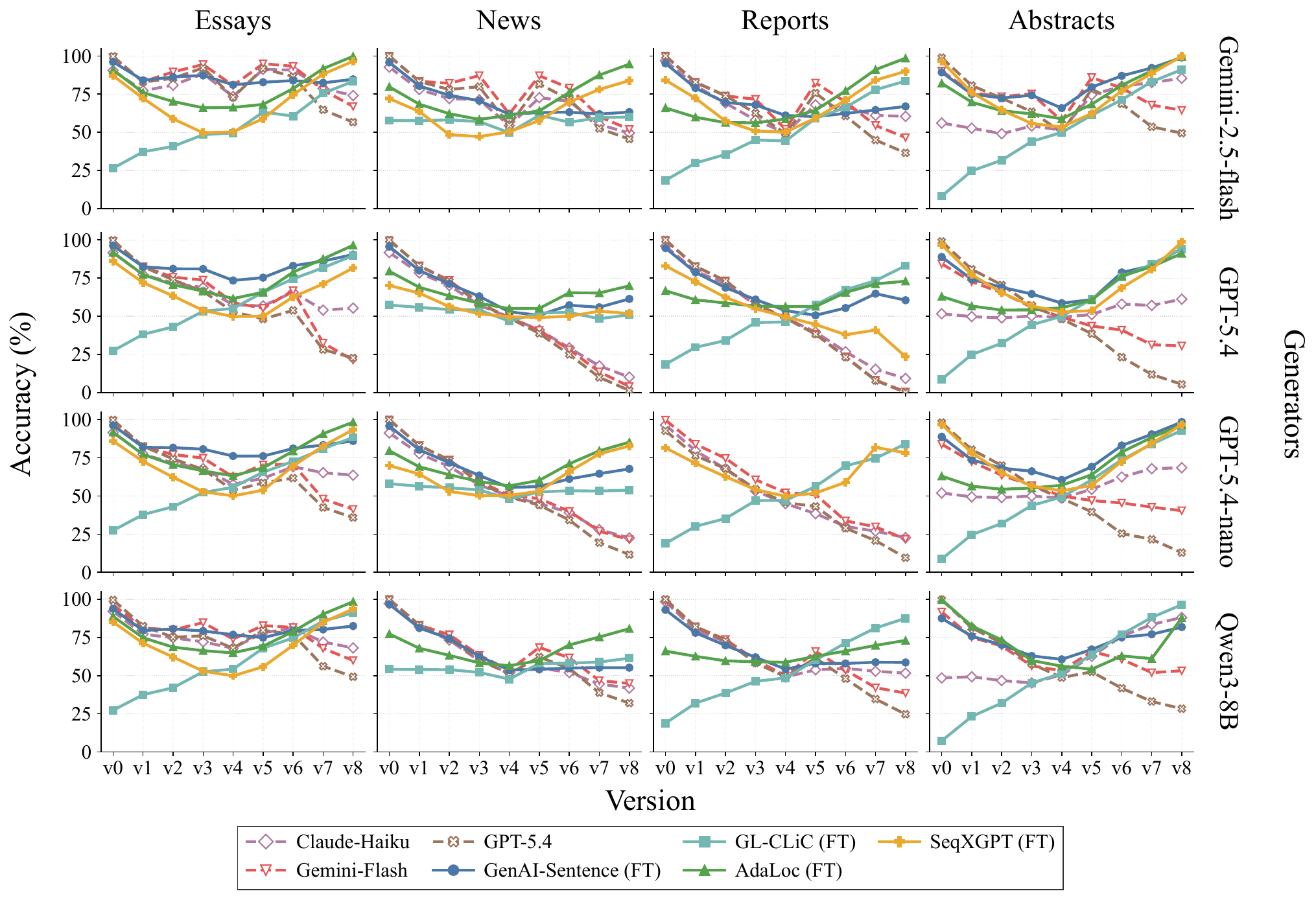}
    \caption{Sentence-level accuracy across revision versions, broken down by domain and generator. The figure compares LLM-as-detectors with fine-tuned sentence-level detectors (FT), showing how detector behavior changes across the progressive human-to-AI revision trajectory.}
    \label{fig:app_sent_acc_1_ft_main}
\end{figure*}

Detection performance does not increase smoothly as AI coverage grows, but instead varies across revision stages, domains, and detector types. Table~\ref{tab:main_table_aggregate} summarizes aggregate results across granularities, averaged over generators. Figures~\ref{fig:app_doc_acc_1_main} and~\ref{fig:app_sent_acc_1_ft_main} show document- and sentence-level accuracy by domain and generator, with full breakdowns in Appendices~\ref{app:results_by_domain_generator} and~\ref{Aggregate_section_appendix}.

\noindent{\bf Document-level detectors show domain-specific failure patterns.}
While aggregate results in Table~\ref{tab:main_table_aggregate} suggest overall trends, Figure~\ref{fig:app_doc_acc_1_main} shows that zero-shot document-level detectors exhibit substantial variation across domains and revision stages, rather than following a consistent trend with increasing AI coverage. RADAR is effective mainly on essays: its F1-AI remains near zero on reports, news, and abstracts across the trajectory, but increases on essays from $50.4$ at $v_1$ to $71.5$ at $v_8$. Fast-DetectGPT shows a different pattern. It detects early edits in reports and news, reaching $61.3$ and $65.0$ F1-AI at $v_1$, but declines as revision progresses, falling to $29.1$ and $35.2$ at $v_8$. These results suggest that document-level detectors rely on domain- and operation-sensitive cues, so increasing AI coverage alone does not guarantee easier detection.

\noindent{\bf Sentence-level supervision improves stability, but does not remove revision effects.}
Sentence-level results show a clear gap between detector families. LLM-as-detectors often improve through early revisions, drop near the compression step, and only partially recover afterward. For example, Gemini-Flash on essays rises from $9.5$ F1-AI at $v_1$ to $71.5$ at $v_3$, drops to $53.6$ at $v_4$, and reaches $57.9$ at $v_8$. Zero-shot sentence-level detectors are less consistent: AdaLoc and GL-CLiC recover in some settings, while SeqXGPT remains near-zero in F1-AI across versions despite high early accuracy. Fine-tuned sentence-level models are generally more stable, with GenAI-Sentence and AdaLoc often recovering at higher AI coverage. However, Figure~\ref{fig:app_sent_acc_1_ft_main} shows that performance still varies across domains and generators, indicating that sentence-level supervision improves robustness but does not eliminate sensitivity to revision stage or edit operation.

\noindent{\bf Fine-grained localization follows a different trade-off.}
DAMASHA operates at the token level and GigaCheck at the span level, so we treat them as localization methods rather than directly comparable document classifiers. Unlike document- and sentence-level detectors, fine-grained methods generally show decreasing accuracy but increasing F1-AI as AI-revision-labeled regions become denser. This reflects class imbalance in early versions: predicting mostly human yields high accuracy when few word-level tokens or spans are labeled AI, while F1-AI becomes more informative at higher AI coverage. Nevertheless, localization remains domain- and generator-dependent, with news consistently harder than essays or abstracts.

Overall, the main trajectory shows that AI-text detection depends on more than the amount of AI-edited content. The ablations in Section~\ref{sec:main_ablation} further disentangle this effect, showing that the degradation around $v_4$ is linked to compression as well as intermediate human--AI coverage.

\subsection{Controlled Analysis of Coverage, Operation, and Revision History}
\label{sec:main_ablation}

The main trajectory varies three factors jointly: AI coverage, edit operation, and cumulative revision history. We therefore run controlled analyses to clarify the non-monotonic patterns in the main results, especially around $v_4$, where intermediate human--AI coverage coincides with compression. Unless otherwise stated, we use \textit{Gemini-2.5-Flash}, since it gives the clearest separation from human-written text in the main experiments.

\noindent{\bf Varying coverage within each operation.}
We first fix the edit operation and vary target AI coverage. Figures~\ref{fig:ablation1_acc_sent} and~\ref{fig:ablation1_f1_sent} show that sentence-level performance does not uniformly collapse at 50\% coverage. Instead, trends differ by operation, with compression generally harder to detect than paraphrase or expansion at comparable coverage. Document-level results show a similar but more detector-dependent pattern (Figures~\ref{fig:ablation1_acc_doc} and~\ref{fig:ablation1_f1_doc}). Thus, the degradation around $v_4$ is not explained by the human--AI ratio alone.

\noindent{\bf Comparing operations at fixed coverage.}
We next fix one coverage level at a time (25\%, 50\%, or 75\%) and compare \textit{compress}, \textit{paraphrase}, and \textit{expand}. Figures~\ref{fig:ablation2_acc_sent} and~\ref{fig:ablation2_f1_sent} show that, for LLM-as-detectors, detectability generally increases from compression to expansion, especially at higher coverage levels. Some zero-shot detectors remain weak or nearly flat across operations. Document-level results show a similar but less uniform trend; full results are provided in Appendix~\ref{app:ablation2}. These results show that edit operation affects detectability even when AI coverage is fixed.

\noindent{\bf Cumulative versus independent editing.}
Finally, we compare the main cumulative trajectory with an independent-editing setting, where each version is edited directly from the original human source. This preserves the same version schedule while removing accumulated re-editing. As shown in Appendix~\ref{app:ablation3}, both sentence- and document-level results remain non-monotonic, including a visible drop around the compression step. The independent trajectories are generally smoother and the drop is less pronounced, indicating that cumulative revision can amplify, but does not fully account for, the observed instability.

\section{Conclusion}

We introduced {\bf OpAI-Bench}, an operation-guided benchmark for evaluating AI-text detection under progressive human--AI revision. Unlike endpoint-based benchmarks, OpAI-Bench preserves intermediate revision states and multi-granularity provenance, enabling detection to be studied as a trajectory rather than a single binary decision.
Our results show that AI-text detectability is not governed by AI coverage alone. Across document, sentence, token, and span granularities, detector behavior depends strongly on edit operation, domain, generator, and detector family. In particular, mixed-authorship intermediate versions can be harder to detect than both human-written and heavily AI-edited endpoints, revealing non-monotonic failure modes that static evaluations miss. Controlled ablations further show that compression is often harder to detect than expansion at matched coverage. These findings suggest that reliable AI-text detection requires moving beyond endpoint human-versus-AI classification toward trajectory-aware and operation-aware evaluation. OpAI-Bench provides a controlled testbed for this setting and highlights the need for detectors that can localize and reason about partial, incremental, and operation-specific AI involvement.

\section*{Acknowledgments}

This work is supported by the United Al Saqer Group Grant.

\bibliographystyle{plainnat}
\bibliography{references}
\clearpage
\appendix

\section*{\Large{Appendix}}

\section{Limitations}
\label{app:limitations}

OpAI-Bench provides a controlled benchmark for studying progressive human-to-AI text transformation, but it still has several limitations. First, the revision trajectories are constructed under predefined AI coverage levels and edit operations, which may not fully capture the diversity and unpredictability of real-world human–AI writing workflows. Second, the benchmark focuses on a fixed set of domains, detectors, and AI editing models, so conclusions may not directly generalize to unseen domains, emerging writing assistants, or future detector architectures. Third, preserving fine-grained authorship provenance requires controlled generation and alignment procedures, which may introduce artifacts that differ from naturally occurring collaborative writing.

\section{Societal Impacts}
\label{app:Social_impact}

This work can have positive societal impacts by improving the transparency and accountability of AI-assisted writing. By analyzing how AI authorship signals evolve under progressive editing, OpAI-Bench can support the development of more reliable detection and provenance tools for education, publishing, journalism, and research integrity. Its multi-granularity annotations may also help move AI-text detection beyond coarse document-level judgments toward more interpretable evidence, reducing the risk of unsupported accusations when only small portions of a text are AI-edited.

The study may also have negative societal impacts. More capable detection benchmarks could be misused to train stronger evasion strategies, enabling users to deliberately rewrite AI-generated content to bypass detectors. Detection tools developed from such benchmarks may also be over-relied upon in high-stakes settings, despite known risks of false positives, domain bias, and uneven performance across writing styles, languages, and user populations. Therefore, OpAI-Bench should be used as an evaluation resource rather than as a definitive authority for authorship judgment, and any deployment of AI-text detection should include human review and clear limitations.

\section{Ethics statement.}
This work follows the NeurIPS Code of Ethics. OpAI-Bench is constructed from cited source datasets and model-generated revisions, and does not involve new human-subject experiments or the collection of private user data. The released benchmark and code are intended to support reproducible research on AI-text detection under progressive editing scenarios.

\section{Detector descriptions}
\label{app:detector_descriptions}

We evaluate the following AI-text detectors.

\begin{enumerate}
\item \textbf{\textit{AdaLoc}}~\citep{zhang-etal-2024-machine} is a sentence-level classifier over RoBERTa-large with a sliding window of three adjacent sentences: every window position emits an AI/human label, and overlapping scores are averaged into one prediction per sentence.
\item \textbf{\textit{RADAR}}~\citep{hu2023radar} is a Vicuna-7B classifier trained adversarially against a paraphraser: the paraphraser produces hard negatives during training, pushing the classifier to learn signals that survive paraphrasing.
\item \textbf{\textit{Fast-DetectGPT}}~\citep{bao2023fast} is zero-shot and scores a candidate by its \emph{conditional probability curvature}: the likelihood of the observed tokens under a scoring LM is compared to the average likelihood of perturbed alternatives drawn from a sampling LM. AI text has higher curvature than human text and the whole score is computed in a single forward pass.
\item \textbf{\textit{DAMASHA}}~\citep{teja-etal-2026-damasha} is a token-level CRF tagger over a dual encoder. RoBERTa-base and ModernBERT-base read the same input; their hidden states are fused by an Info-Mask layer driven by simple stylistic features, and the CRF decodes the per-token AI/human tag sequence.
\item \textbf{\textit{GigaCheck}}~\citep{tolstykh2024gigacheck} is a DETR-style span detector on top of Mistral-7B: the LM encodes tokens and a DETR decoder predicts a fixed-size set of character intervals, each labelled AI or human, alongside a coarse document-level head.
\item \textbf{\textit{Desklib}}~\citep{desklib_ai_text_detector_v101} is a single-transformer document-level classifier; we use the public Hugging Face release out of the box, with no further training.
\item \textbf{\textit{E5-small}}~\citep{wang2022text} is an E5-small encoder with a LoRA adapter trained for AI-text classification by the original authors. We use the public weights directly as a document-level binary classifier.
\item \textbf{\textit{OOD-LLM-Detect}}~\citep{zeng2025human} treats AI-text detection as one-class classification: a Deep~SVDD model is fitted to language-model embeddings of human text only, and a candidate is scored by its distance from the learnt human-text region.
\item \textbf{\textit{RoBERTa-OpenAI}}~\citep{solaiman2019release} is RoBERTa-base fine-tuned by OpenAI on GPT-2 outputs; we use the released document-level binary classifier as is.
\item \textbf{\textit{DetectLLM}}~\citep{su2023detectllm} is zero-shot and combines two ranking statistics under a single reference causal LM: the log-rank ratio (LRR) and the normalised perturbation rank (NPR). Both capture how unusually high-ranked the observed tokens are under the reference distribution.
\item \textbf{\textit{GL-CLiC}}~\citep{adi-etal-2025-gl} is a sentence-level classifier whose feature vector concatenates a DeBERTa contextual embedding, per-sentence global--local coherence scores, and per-sentence lexical complexity statistics; the resulting features are passed through a small classification head.
\item \textbf{\textit{SeqXGPT}}~\citep{wang-etal-2023-seqxgpt} represents each token by its log-probability under four reference LMs (\texttt{gpt2-xl}, \texttt{gpt-neo-2.7B}, \texttt{gpt-j-6B}, \texttt{llama-7B}), yielding a $(T,\,4)$ feature matrix. A small CNN\,+\,Transformer\,+\,CRF stack reads this matrix and emits per-word labels, which are aggregated to sentence level.
\item \textbf{\textit{GPT-5.4}} (reasoning level: none) is an API-based judge prompted with the candidate document and asked to return a per-sentence AI/human label list directly.
\item \textbf{\textit{Gemini 3 Flash}} (thinking level: minimal) is an API-based judge prompted with the candidate document and asked to return a per-sentence AI/human label list directly.
\item \textbf{\textit{Claude Haiku 4.5}} (reasoning level: minimal) is an API-based judge prompted with the candidate document and asked to return a per-sentence AI/human label list directly.
\item \textbf{\textit{GenAI-Sentence}}~\citep{teja2025fine} is a token-level CRF tagger: a DeBERTa backbone feeds a BiGRU encoder, a linear classifier, and a CRF decoder that emits per-token AI/human labels; sentence labels are obtained by aggregation.
\end{enumerate}

\section{Implementation Details}
\label{app:implementation}

\subsection{Text Normalization and Segmentation}
\label{app:segmentation}

All source documents are normalized prior to processing: line endings are standardized, consecutive blank lines are collapsed to a single paragraph boundary, and leading and trailing whitespace is removed. Documents are segmented into paragraphs by splitting on blank lines, and each paragraph is assigned a stable identifier preserved across all derived versions. Sentence segmentation is performed using NLTK's \texttt{sent\_tokenize}, applied independently to each paragraph. Sentences receive stable identifiers assigned sequentially across paragraphs, used for sentence selection, provenance tracking, and document reconstruction across all nine versions.

\subsection{Deterministic Sentence Selection}
\label{app:selection}

For each source document $D$ with $N$ sentences, a single fixed ordering $\pi(D) = (\pi_1, \ldots, \pi_N)$ is computed once and reused across all versions. The shuffle seed is derived deterministically from the document identifier, ensuring reproducibility and content-agnostic ordering. Given target coverage $c^{(t)}_\text{sent} \in [0,1]$ at version $t$, the number of sentences selected is:

\begin{equation}
    k^{(t)} = \left\lceil c^{(t)}_\text{sent} \cdot N 
    \right\rceil.
\end{equation}

The edited set is constructed cumulatively: previously selected sentences are always retained, and new sentences are drawn from $\pi(D)$ to reach $k^{(t)}$, guaranteeing $S^{(0)} \subseteq S^{(1)} \subseteq \cdots \subseteq S^{(8)}$.

\subsection{Token- and Span-Level Provenance}
\label{app:provenance}

Token- and span-level provenance labels are derived deterministically from the editing process, not from 
post-hoc alignment or model-provided markers, ensuring annotations are exact and reproducible.

\noindent{\bf Span derivation.}
During construction, each sentence selected for AI rewriting is wrapped in XML-style delimiters before being passed to the LLM, and the returned rewrite is stored with those delimiters intact. After each version is constructed, the tagged document $D^{(t)}_\text{tagged}$ is parsed to extract character-level 
AI spans $\mathcal{A}^{(t)}_\text{char} = \{(a_\ell, b_\ell)\}_{\ell=1}^{L}$, where $a_\ell$ and $b_\ell$ are character offsets in the clean text $D^{(t)}$. Span boundaries reflect precisely which characters were produced by the LLM rewriting step.

\noindent{\bf Word-level projection.}
Word-level labels are projected from character spans via whitespace tokenization. For word $w_j$ with character span $[s_j, e_j]$:

\begin{equation}
    y^{(t)}_j = \mathbf{1}\!\left[
        \exists\,(a_\ell, b_\ell) \in 
        \mathcal{A}^{(t)}_\text{char}
        \;\text{s.t.}\;
        \max(s_j, a_\ell) < \min(e_j, b_\ell)
    \right].
    \label{eq:word_label}
\end{equation}

Consecutive AI-labeled words are merged into contiguous spans. Because the editing unit is the sentence, each AI span corresponds to one rewritten sentence; boundaries therefore reflect sentence boundaries rather than intra-sentence partial rewrites.

\section{Benchmark Statistics}

\begin{table}[H]
\centering
\resizebox{1\linewidth}{!}{  
\begin{tabular}{l l c c c c}
\toprule
\textbf{Domain} & \textbf{Generator} & \textbf{\# Traj.} & \textbf{\# Versioned Samples} & \textbf{Avg.\ sents} & \textbf{Avg.\ tokens} \\
\midrule
\multirow{3}{*}{Student essays}
  & GPT-5.4 & 1,969 & 17,721 & 20.9 & 383.6 \\
  & GPT-5.4-nano & 1,968 & 17,712 & 20.9 & 397.1 \\
  & Gemini-2.5-flash & 3,969 & 35,721 & 21.0 & 407.2 \\
\midrule
\multirow{3}{*}{News articles}
  & GPT-5.4 & 1,904 & 17,136 & 24.0 & 477.0 \\
  & GPT-5.4-nano & 1,998 & 17,982 & 24.0 & 479.4 \\
  & Gemini-2.5-flash & 3,990 & 35,910 & 23.9 & 504.1 \\
\midrule
\multirow{3}{*}{Government reports}
  & GPT-5.4 & 2,000 & 18,000 & 20.5 & 539.4 \\
  & GPT-5.4-nano & 2,000 & 18,000 & 20.6 & 552.5 \\
  & Gemini-2.5-flash & 4,000 & 36,000 & 20.6 & 581.4 \\
\midrule
\multirow{3}{*}{Scientific abstracts}
  & GPT-5.4 & 1,764 & 15,876 & 11.0 & 224.1 \\
  & GPT-5.4-nano & 1,764 & 15,876 & 11.0 & 227.7 \\
  & Gemini-2.5-flash & 3,763 & 33,860 & 11.0 & 242.2 \\
\midrule
\textbf{Total} & & 31,089 & 279,794 & 19.3 & 426.1 \\
\bottomrule
\end{tabular}
}
\vspace{0.1in}
\caption{Main split statistics by domain and generator. Trajectories are generator-specific revision sequences initialized from human-written \texttt{v0} texts, and versioned samples count all versions from \texttt{v0} to \texttt{v8}.}
\label{tab:dataset_stats_multirow}
\end{table}

\begin{table}[H]
\centering
\small
\begin{tabular}{l r r r r}
\toprule
\textbf{Domain} & \textbf{Train} & \textbf{Development} & \textbf{Test} & \textbf{Total} \\
\midrule
Student essays & 49{,}149 & 11{,}205 & 10{,}800 & 71{,}154 \\
News articles & 50{,}688 & 10{,}566 & 9{,}774 & 71{,}028 \\
Government reports & 50{,}049 & 10{,}881 & 11{,}070 & 72{,}000 \\
Scientific abstracts & 46{,}145 & 9{,}711 & 9{,}756 & 65{,}612 \\
\midrule
\textbf{Total} & 196{,}031 & 42{,}363 & 41{,}400 & 279{,}794 \\
\bottomrule
\end{tabular}
\vspace{0.1in}
\caption{Train, development, and test row counts for the primary-generator benchmark subset, broken down by domain. Counts are reported over GPT-5.4, GPT-5.4-nano, and Gemini-2.5-Flash.}
\label{tab:dataset_split_stats}
\end{table}

\subsection{Detailed Results by Domain and Generator}
\label{app:results_by_domain_generator}

This appendix provides the full per-domain and per-generator results that complement the aggregate analyses in the main text. For each evaluation granularity, we report both accuracy and F1-AI. Accuracy captures overall classification behavior, while F1-AI focuses on the detector's ability to identify AI-edited content, which is especially important under mixed-authorship settings where the class distribution changes across versions.

\subsubsection{Document-Level Results}
\label{Acc_F1_doc_figure}

\begin{figure*}[!htb]
    \centering
    \includegraphics[width=\textwidth]{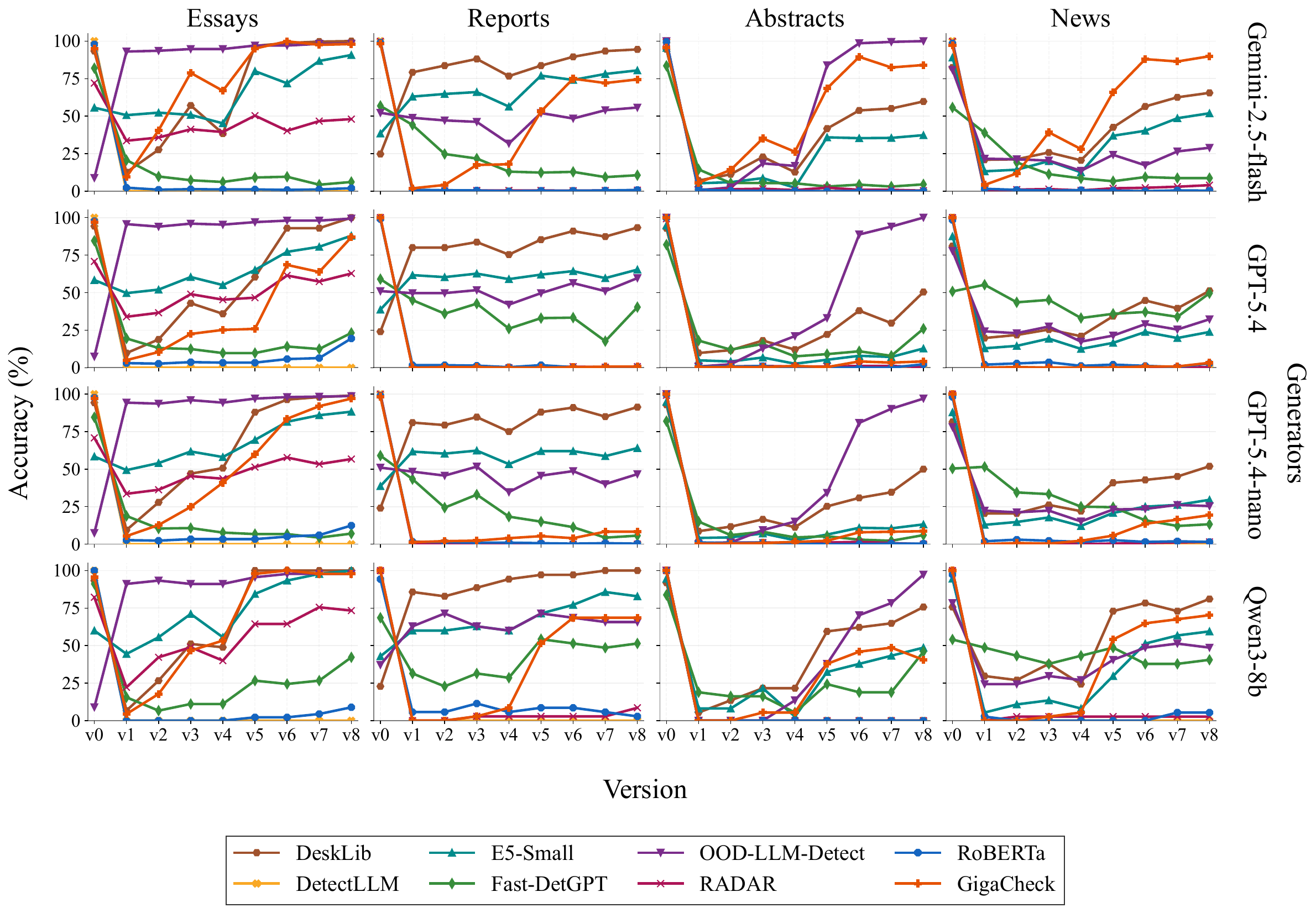}
    \caption{Document-level accuracy broken down by domain and generator.}
    \label{fig:app_doc_acc_1}
\end{figure*}

\begin{figure*}[t]
    \centering
    \includegraphics[width=\textwidth]{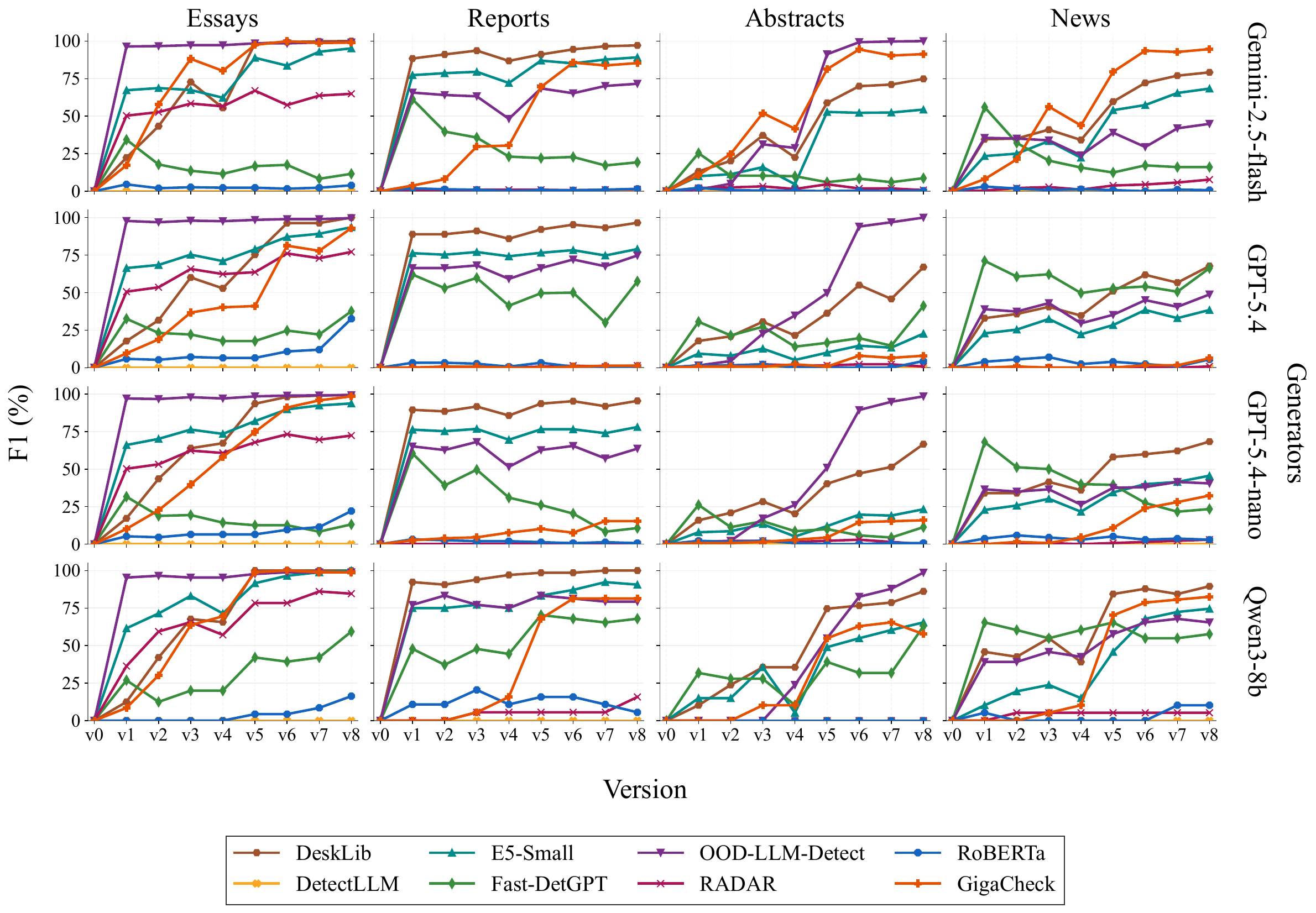}
\caption{Document-level F1-AI across revision versions, domains, and generators. Each curve represents a document-level detector, showing how AI-targeted detection performance changes along the progressive human-to-AI revision trajectory.}
    \label{fig:app_doc_f1}
\end{figure*}

Figures~\ref{fig:app_doc_acc_1} and~\ref{fig:app_doc_f1} report document-level accuracy and F1-AI, respectively, broken down by domain and generator. These figures show how zero-shot document-level detectors behave across the full revision trajectory. They also expose detector-specific and domain-specific variation that is hidden in the aggregate table, including cases where recovery at high AI coverage occurs only for particular generator--domain combinations.
\subsubsection{Sentence-Level Results}

\begin{figure*}[!htb]
    \centering
    \includegraphics[width=\textwidth]{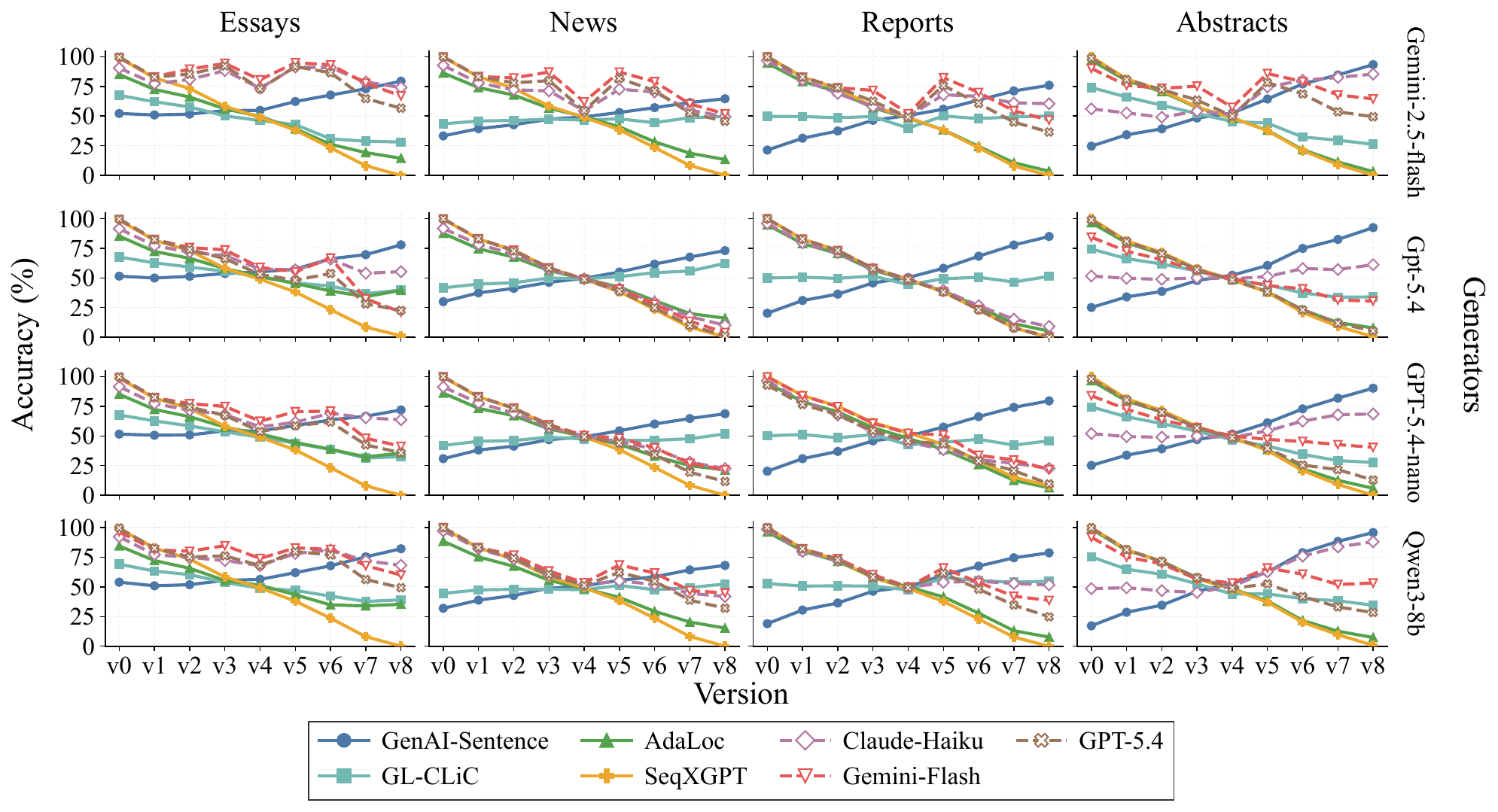}
\caption{Sentence-level accuracy across revision versions, domains, and generators. The figure compares sentence-level detectors across the full revision trajectory, including zero-shot methods and LLM-as-detectors sentence-level models.}
    \label{fig:app_sent_acc_1}
\end{figure*}

\begin{figure*}[!htb]
    \centering
    \includegraphics[width=\textwidth]{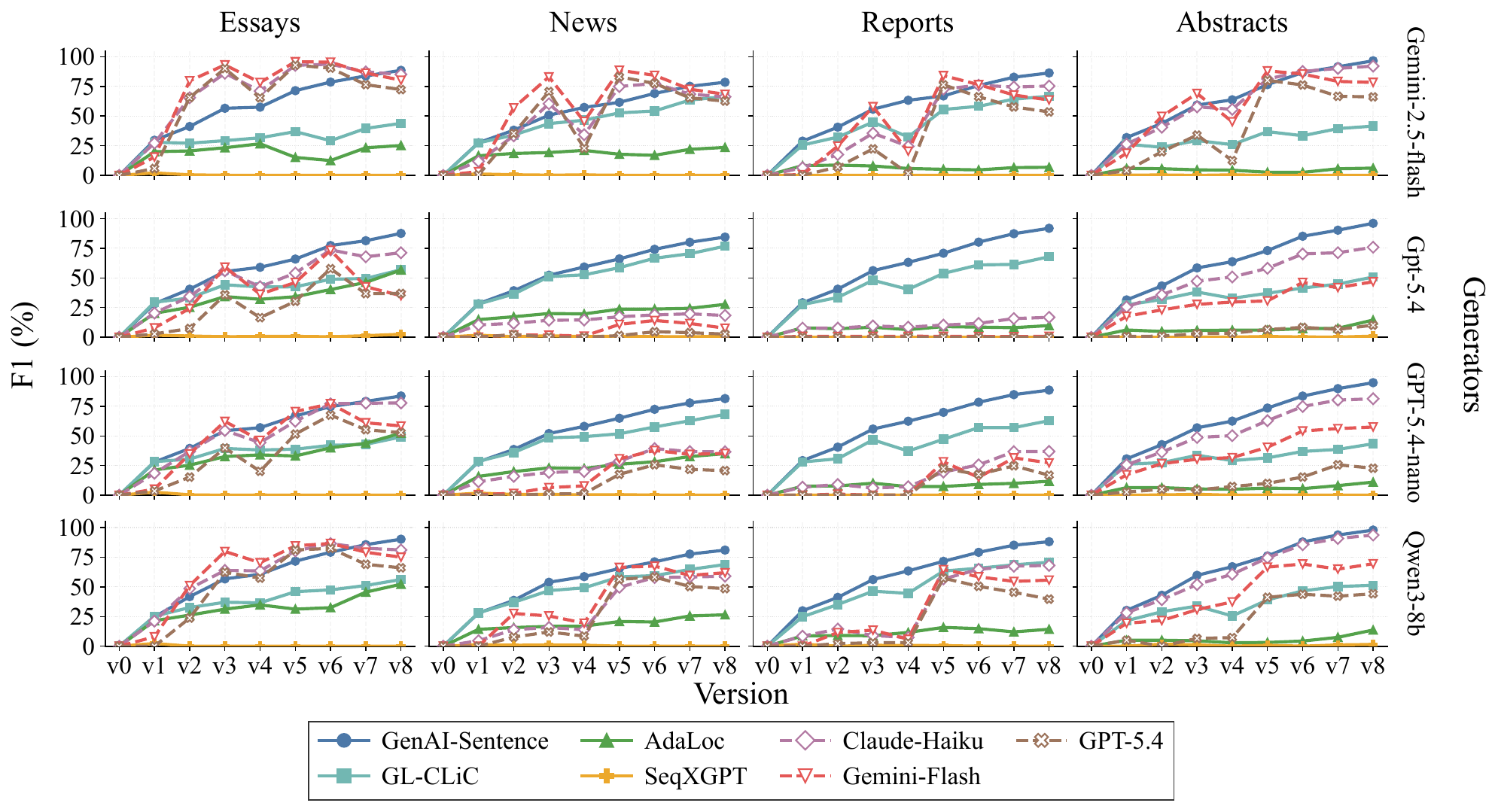}
\caption{Sentence-level F1-AI across revision versions, domains, and generators. The figure compares sentence-level detectors across the full revision trajectory, including zero-shot methods and LLM-as-detectors.}
    \label{fig:app_sent_f1}
\end{figure*}

\begin{figure*}[!htb]
    \centering
    \includegraphics[width=\textwidth]{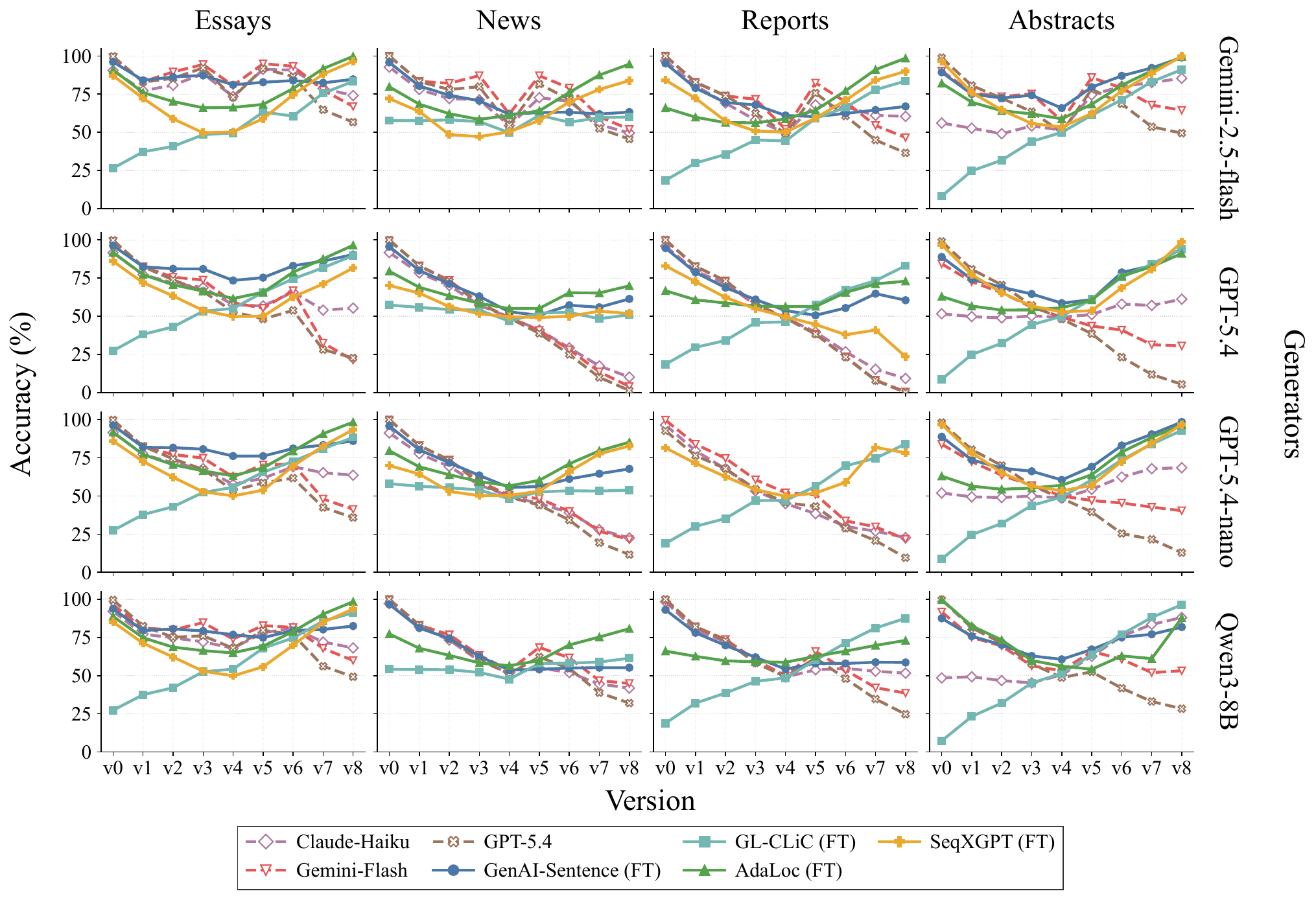}
\caption{Sentence-level accuracy across revision versions, domains, and generators, including fine-tuned sentence-level models. The figure highlights how fine-tuned detectors compare with LLM-as-detector methods along the progressive human-to-AI revision trajectory.}
    \label{fig:app_sent_acc_ft}
\end{figure*}

\begin{figure*}[!htb]
    \centering
    \includegraphics[width=\textwidth]{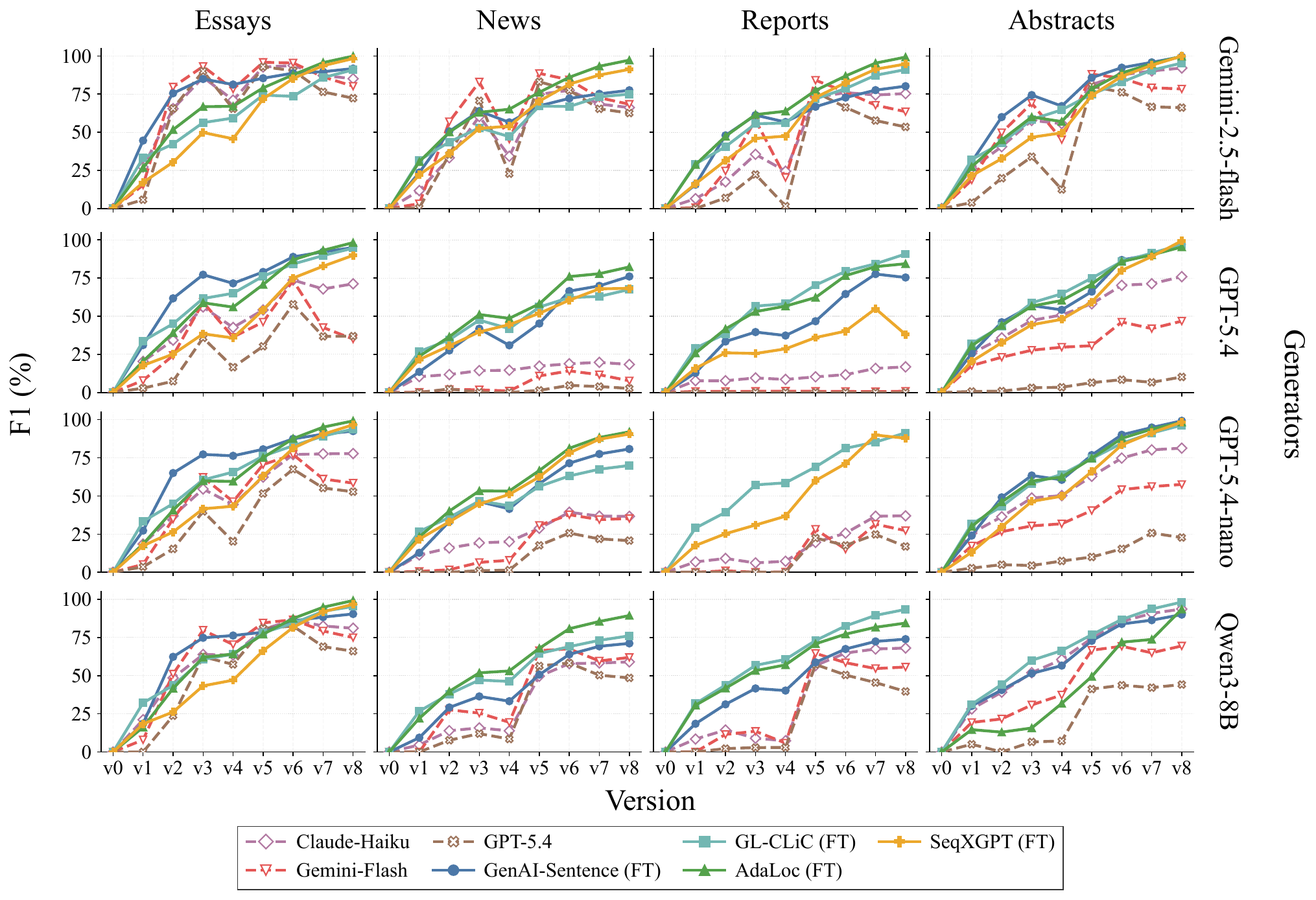}
    \caption{Sentence-level F1-AI across revision versions, domains, and generators, including fine-tuned sentence-level models. The figure highlights how fine-tuned detectors compare with LLM-as-detector methods along the progressive human-to-AI revision trajectory.}
    \label{fig:app_sent_f1_ft}
\end{figure*}

Figures~\ref{fig:app_sent_acc_1} and~\ref{fig:app_sent_f1} report sentence-level accuracy and F1-AI for zero-shot and LLM-as-detector methods. Figures~\ref{fig:app_sent_acc_ft} and~\ref{fig:app_sent_f1_ft} additionally include fine-tuned sentence-level detectors. These breakdowns allow us to compare how detector families respond to progressive rewriting across domains and generators, and to assess whether sentence-level supervision improves stability under mixed-authorship revisions.

\subsubsection{Token- and Span-Level Results}

\begin{figure*}[!htb]
    \centering
    \includegraphics[width=\textwidth]{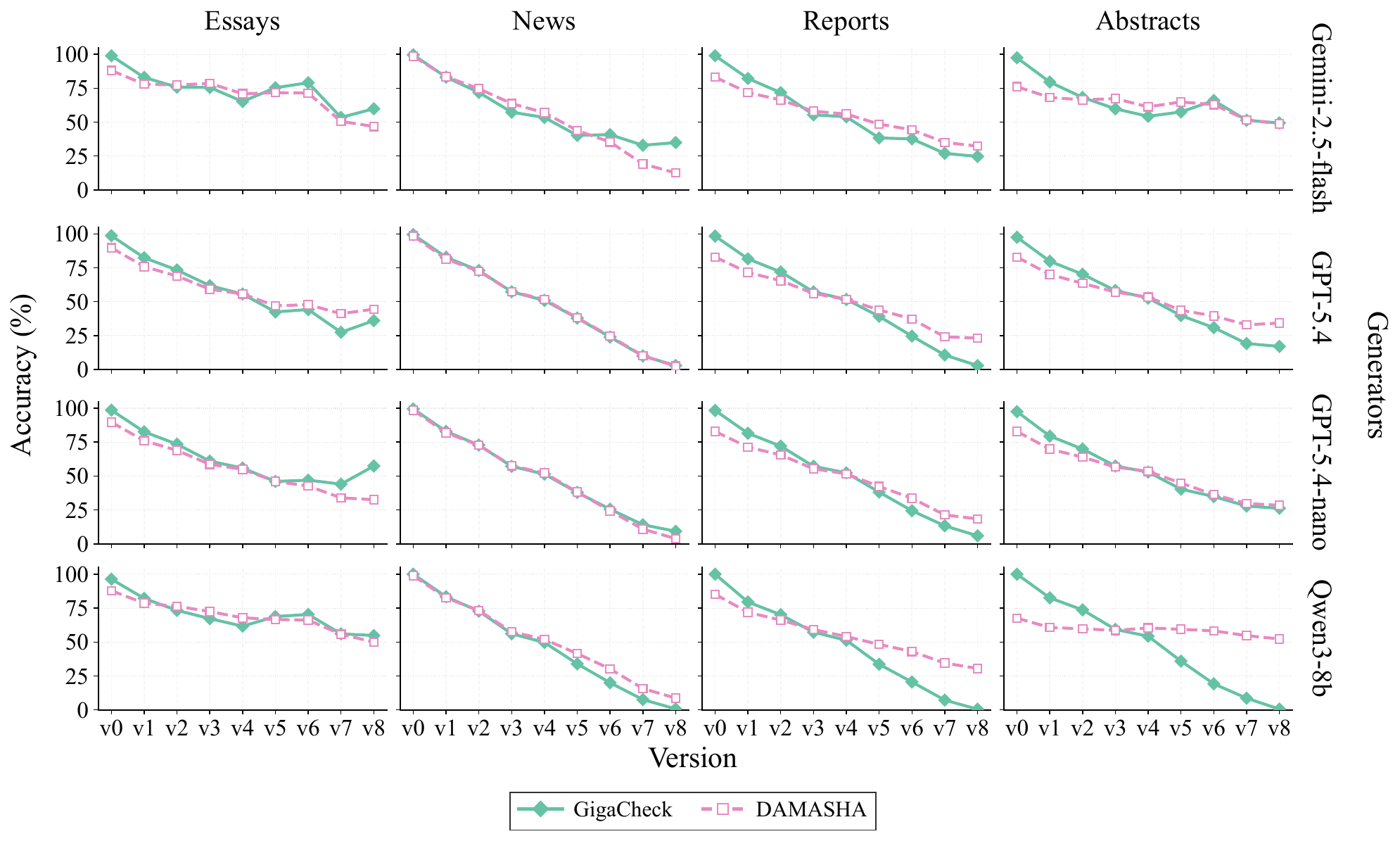}
    \caption{Token- and span-level accuracy broken down by domain and generator.}
    \label{fig:app_token_acc_1}
\end{figure*}

\begin{figure*}[!htb]
    \centering
    \includegraphics[width=\textwidth]{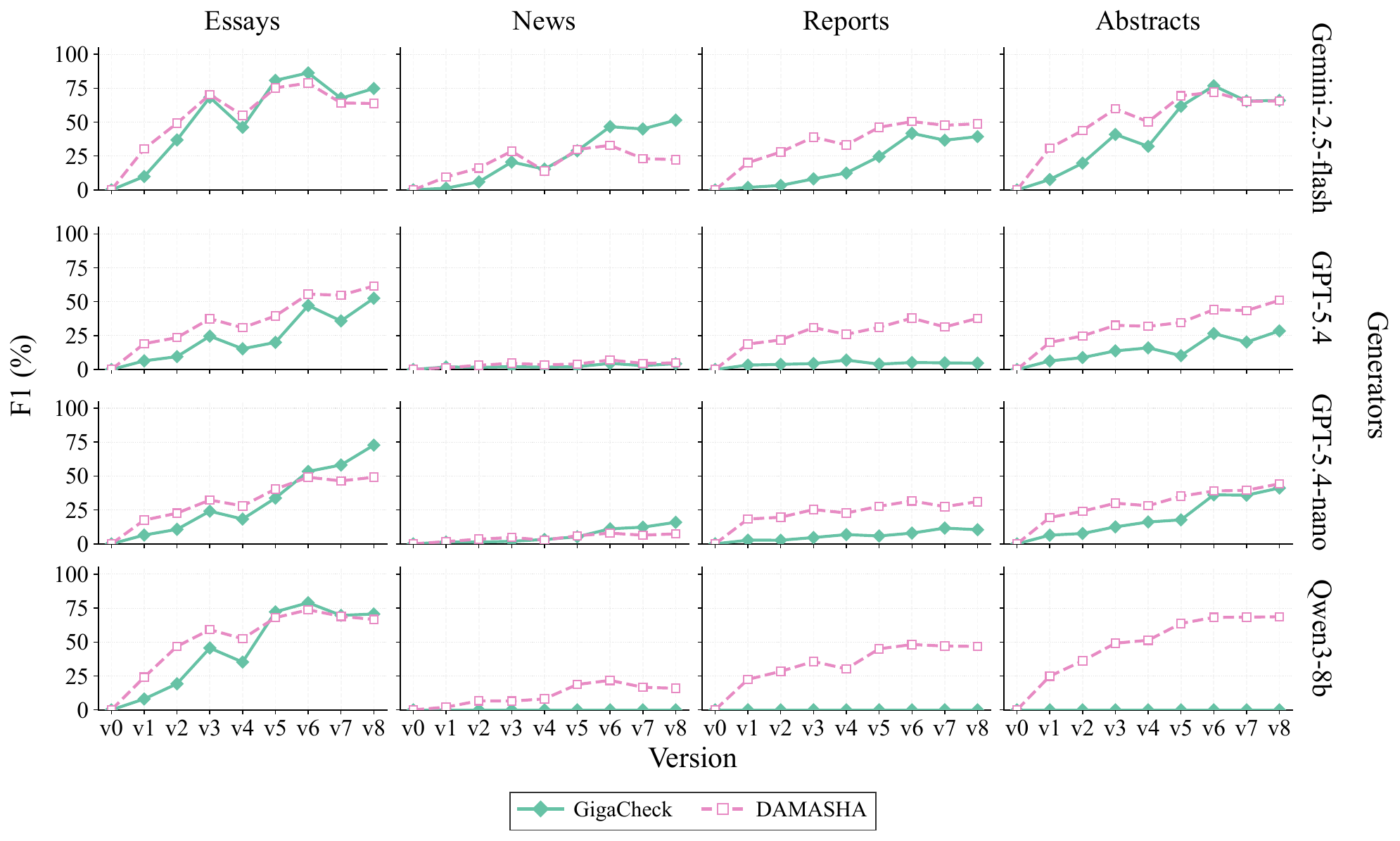}
    \caption{Token- and span-level F1-AI broken down by domain and generator.}
    \label{fig:app_token_f1}
\end{figure*}

Figures~\ref{fig:app_token_acc_1} and~\ref{fig:app_token_f1} report fine-grained token/span-level performance by domain and generator. DAMASHA operates at the token level, while GigaCheck produces span-level predictions. These results complement the coarser document- and sentence-level analyses by showing how localization performance changes as AI-marked regions become denser across the revision trajectory.

\section{Aggregated Results}
\label{Aggregate_section_appendix}

This appendix provides extended aggregate tables averaged over generators. Unlike Appendix~\ref{app:results_by_domain_generator}, which shows full domain--generator breakdowns, the tables below aggregate over generators and report results by domain and revision version. These tables provide a compact view of detector behavior across the full trajectory and include additional metrics beyond those shown in the main text.

\subsection{Sentence Level}
\label{Sent_aggregate_appendix}

Table~\ref{tab:sentence_all_detectors_domains_avg_generators_extended_accuu} reports sentence-level accuracy and F1-AI for all sentence-level detectors, averaged over generators. The table includes zero-shot detectors, LLM-as-detectors, and fine-tuned sentence-level models, enabling direct comparison across detector families. Table~\ref{tab:sentence_all_detectors_domains_avg_generators_extended}  provides the corresponding extended sentence-level metrics, including Macro-F1 and false negative rate. These additional metrics help characterize whether detectors fail by missing AI-edited content or by over-predicting the AI class.

\subsection{Token and Span Level}
 \label{Token_span_aggregate_appendix}

Table~\ref{tab:finegrained_token_span_domains_avg_generators} reports fine-grained token/span-level accuracy and F1-AI, averaged over generators. Table~\ref{tab:finegrained_token_span_domains_avg_generators2} provides the corresponding extended metrics. DAMASHA is evaluated at the token level, while GigaCheck is evaluated at the span level. These results complement the document- and sentence-level tables by evaluating whether detectors can localize AI-edited regions rather than only assign a global label. Because early revision versions contain relatively few AI-marked words or spans, accuracy can remain high even when the detector misses AI-edited regions. F1-AI is therefore particularly important for interpreting fine-grained performance, as it directly measures recovery of the AI-labeled class. The extended metrics further help distinguish between detectors that remain conservative and miss AI edits, and detectors that recover more AI-marked regions as coverage increases.

\section{Ablations}
\label{sec:total_ablation}

We provide three controlled ablations to isolate the factors that vary jointly in the main trajectory. Ablation 1 fixes the edit operation and varies AI coverage, testing whether detector behavior is driven by coverage alone. Ablation 2 fixes AI coverage and varies the edit operation, testing whether different operations remain distinguishable at matched coverage. Ablation 3 compares cumulative editing with independent editing from the original source, testing the effect of accumulated revision history.

\subsection{Ablation 1: Coverage-controlled Edit Operations}
\label{ablation1}

\begin{figure*}[!htb]
    \centering
    \includegraphics[width=\textwidth]{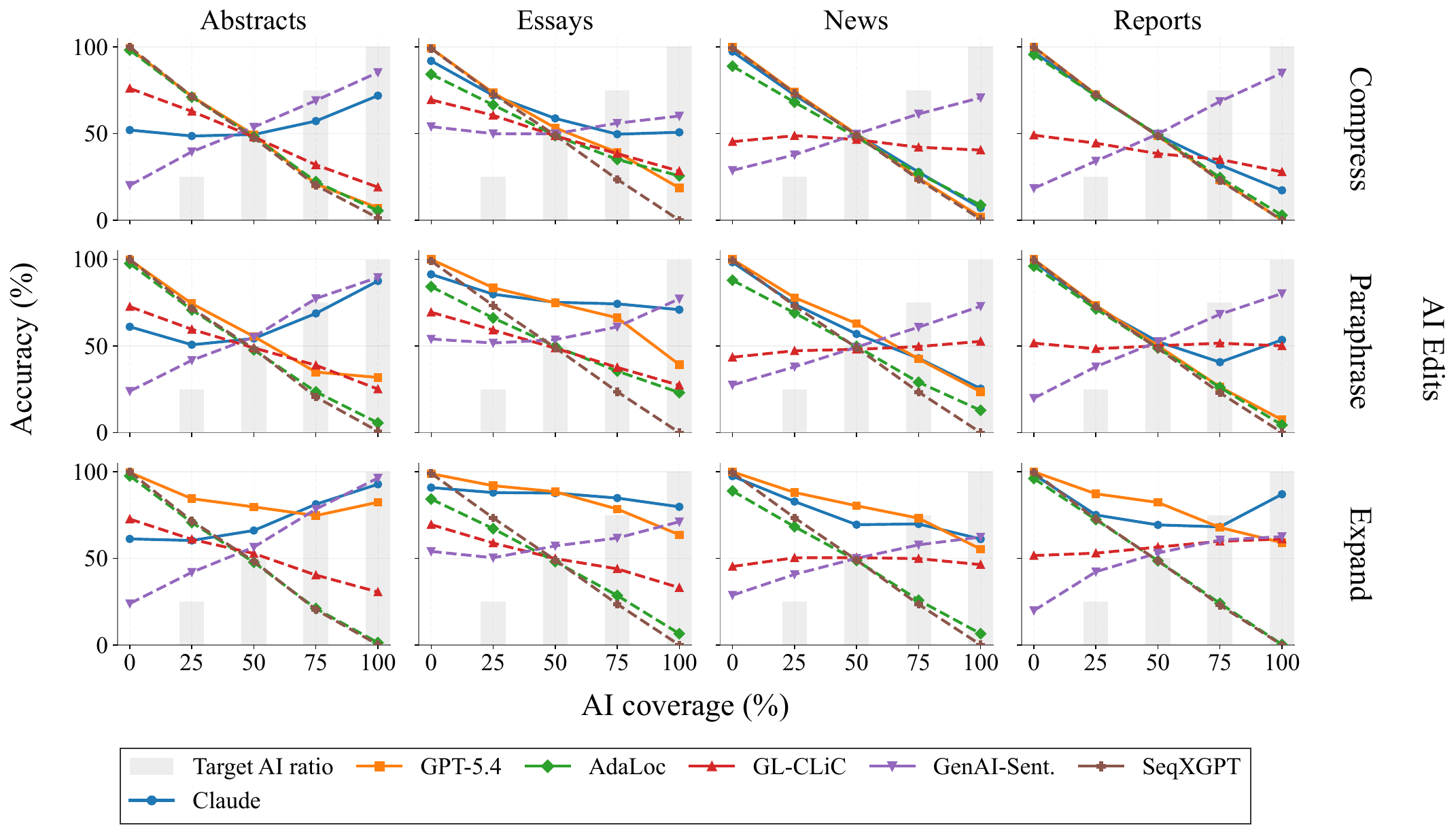}
    \caption{Sentence-level accuracy under coverage-controlled edit operations.}
    \label{fig:ablation1_acc_sent}
\end{figure*}

\begin{figure*}[!htb]
    \centering
    \includegraphics[width=\textwidth]{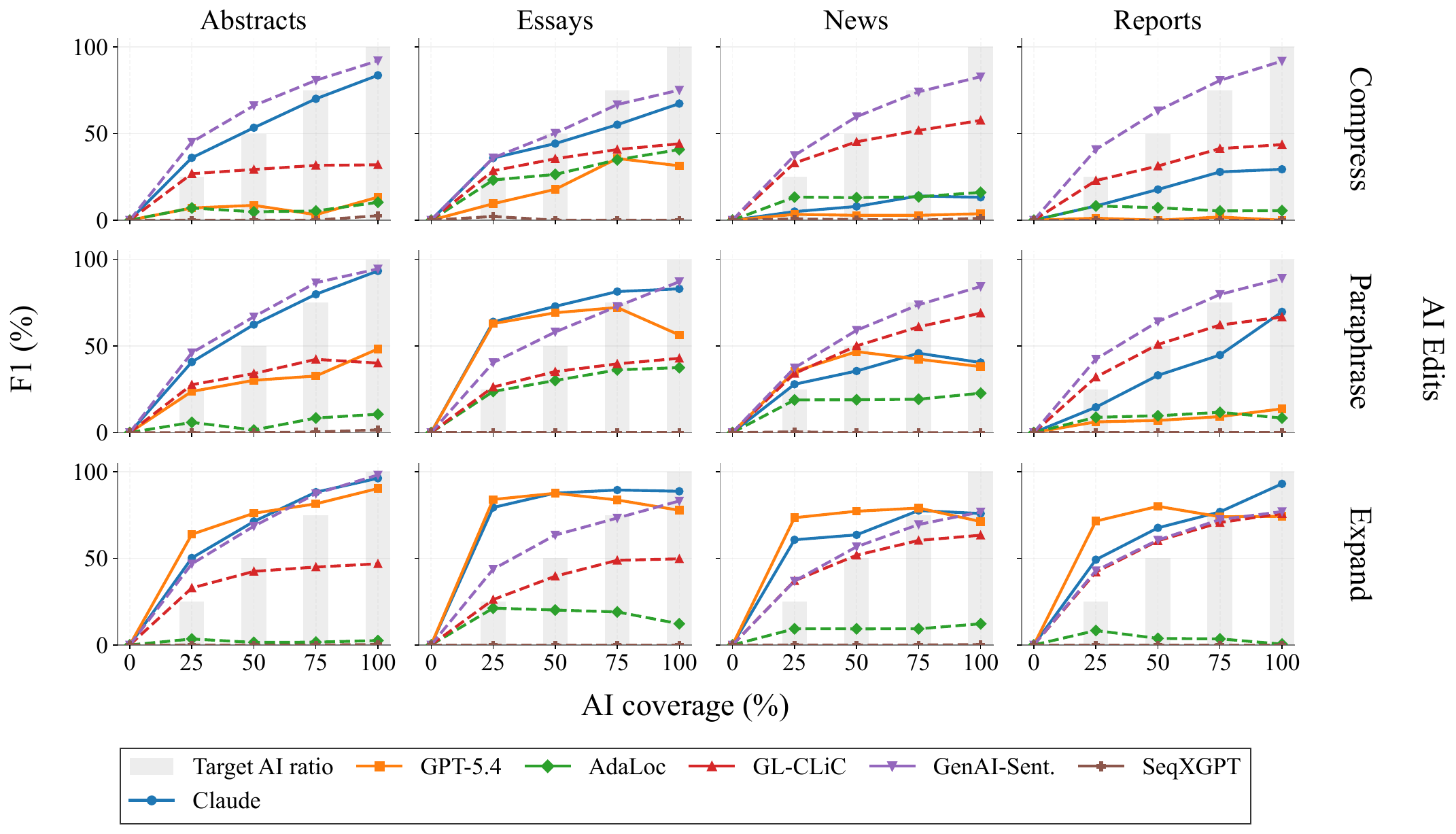}
    \caption{Sentence-level F1-AI under coverage-controlled edit operations.}
    \label{fig:ablation1_f1_sent}
\end{figure*}

\begin{figure*}[!htb]
    \centering
    \includegraphics[width=\textwidth]{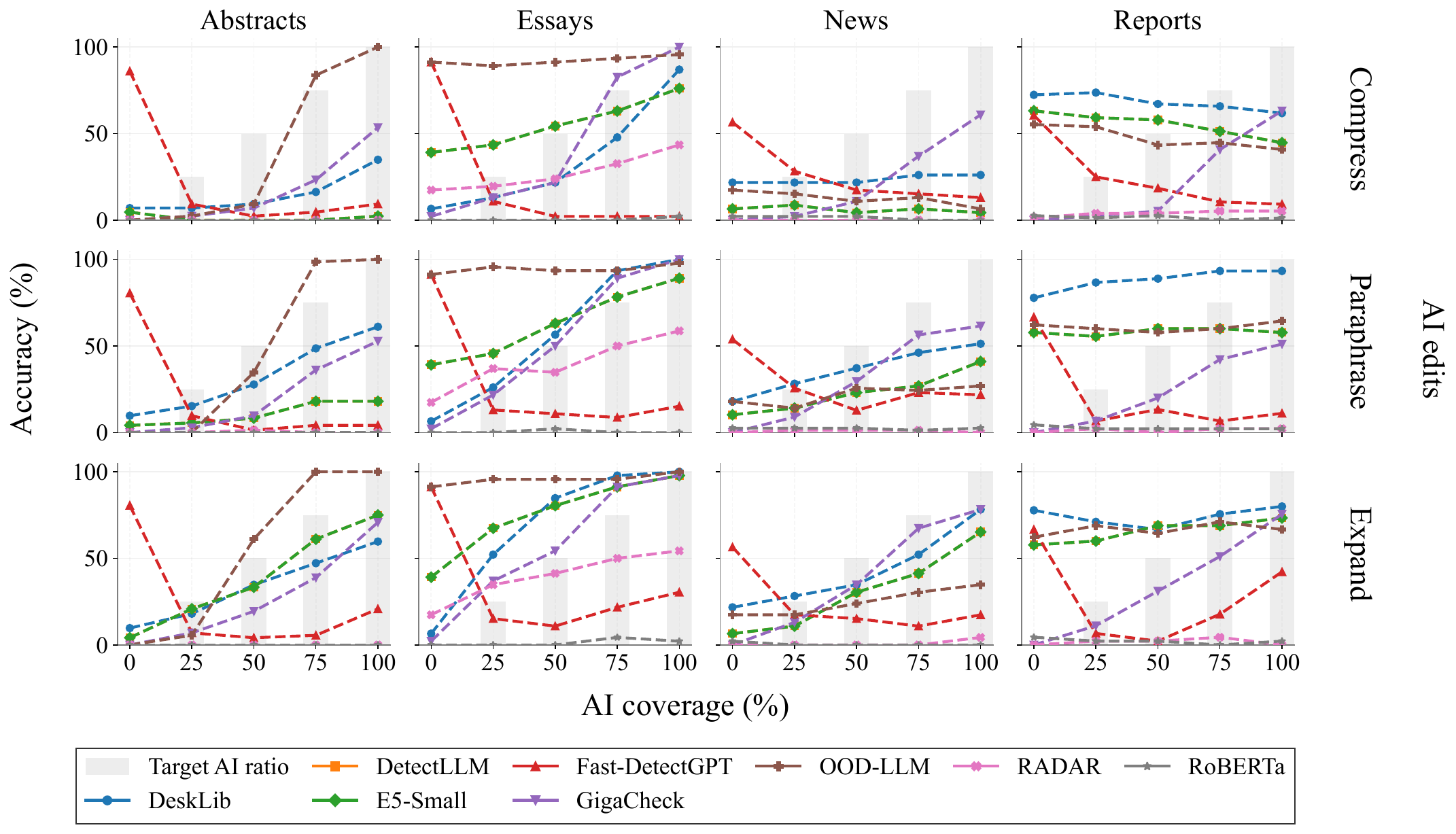}
    \caption{Document-level accuracy under coverage-controlled edit operations.}
    \label{fig:ablation1_acc_doc}
\end{figure*}

\begin{figure*}[!htb]
    \centering
    \includegraphics[width=\textwidth]{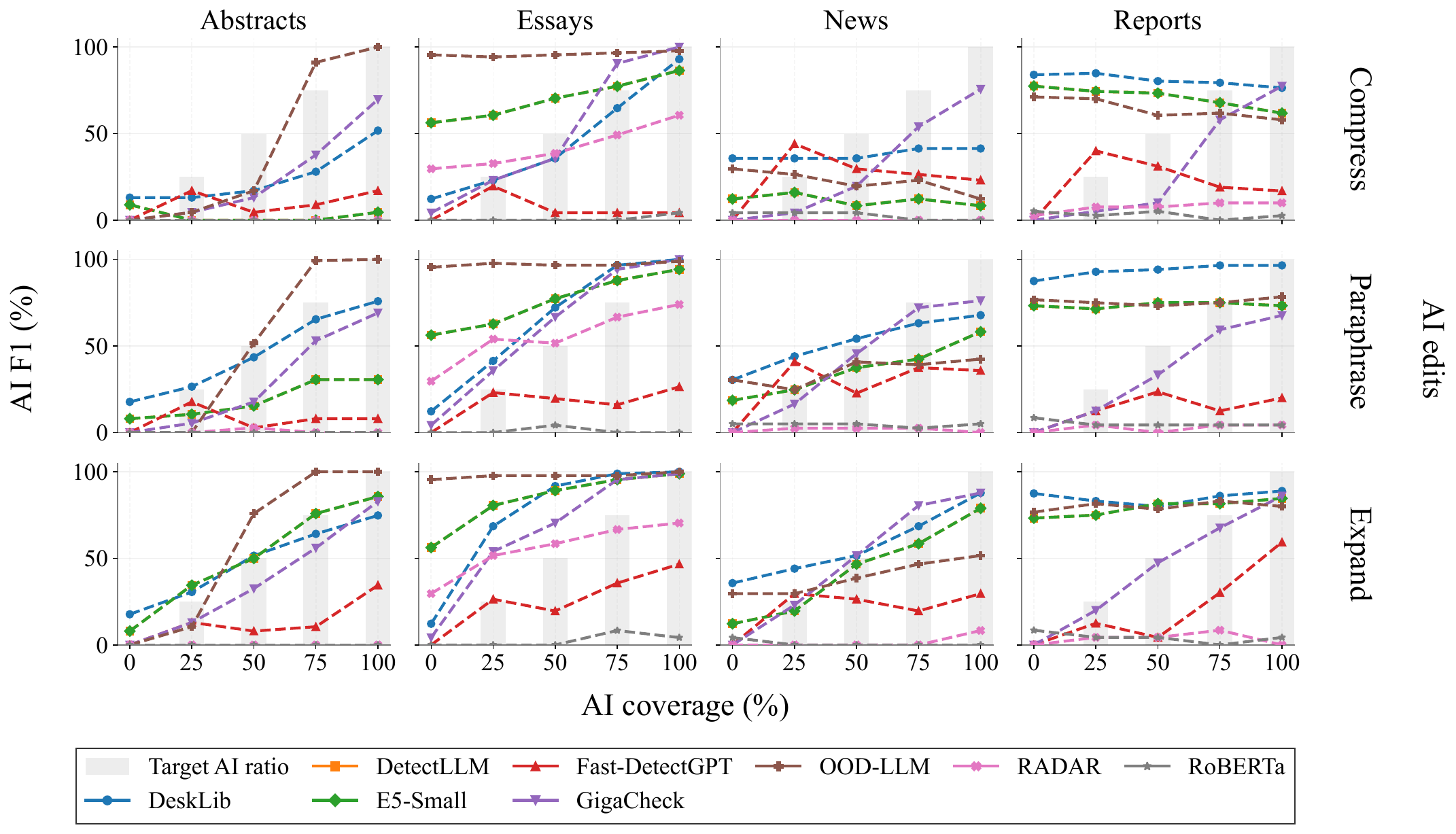}
    \caption{Document-level F1-AI under coverage-controlled edit operations.}
    \label{fig:ablation1_f1_doc}
\end{figure*}

This ablation fixes the edit operation and varies the target AI coverage. Sentence-level accuracy and F1-AI are shown in Figures~\ref{fig:ablation1_acc_sent} and~\ref{fig:ablation1_f1_sent}, respectively. Document-level accuracy and F1-AI are shown in Figures~\ref{fig:ablation1_acc_doc} and~\ref{fig:ablation1_f1_doc}, respectively.

\subsection{Ablation 2: Fixed-coverage Edit Operations}
\label{app:ablation2}

\begin{figure*}[!htb]
    \centering
    \includegraphics[width=\textwidth]{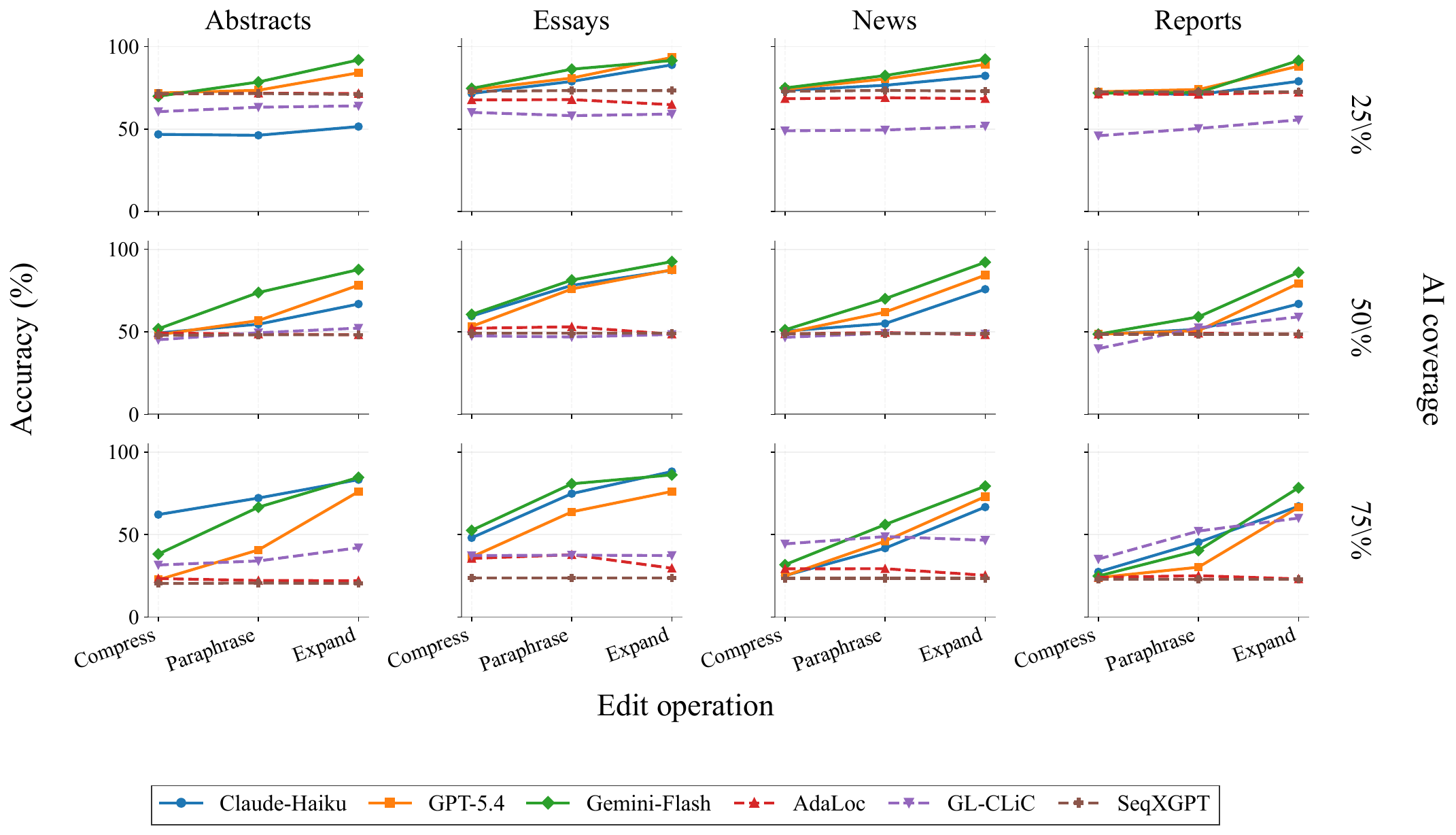}
     \caption{Sentence-level accuracy at fixed AI coverage while varying edit operation.}
    \label{fig:ablation2_acc_sent}
\end{figure*}

\begin{figure*}[!htb]
    \centering
    \includegraphics[width=\textwidth]{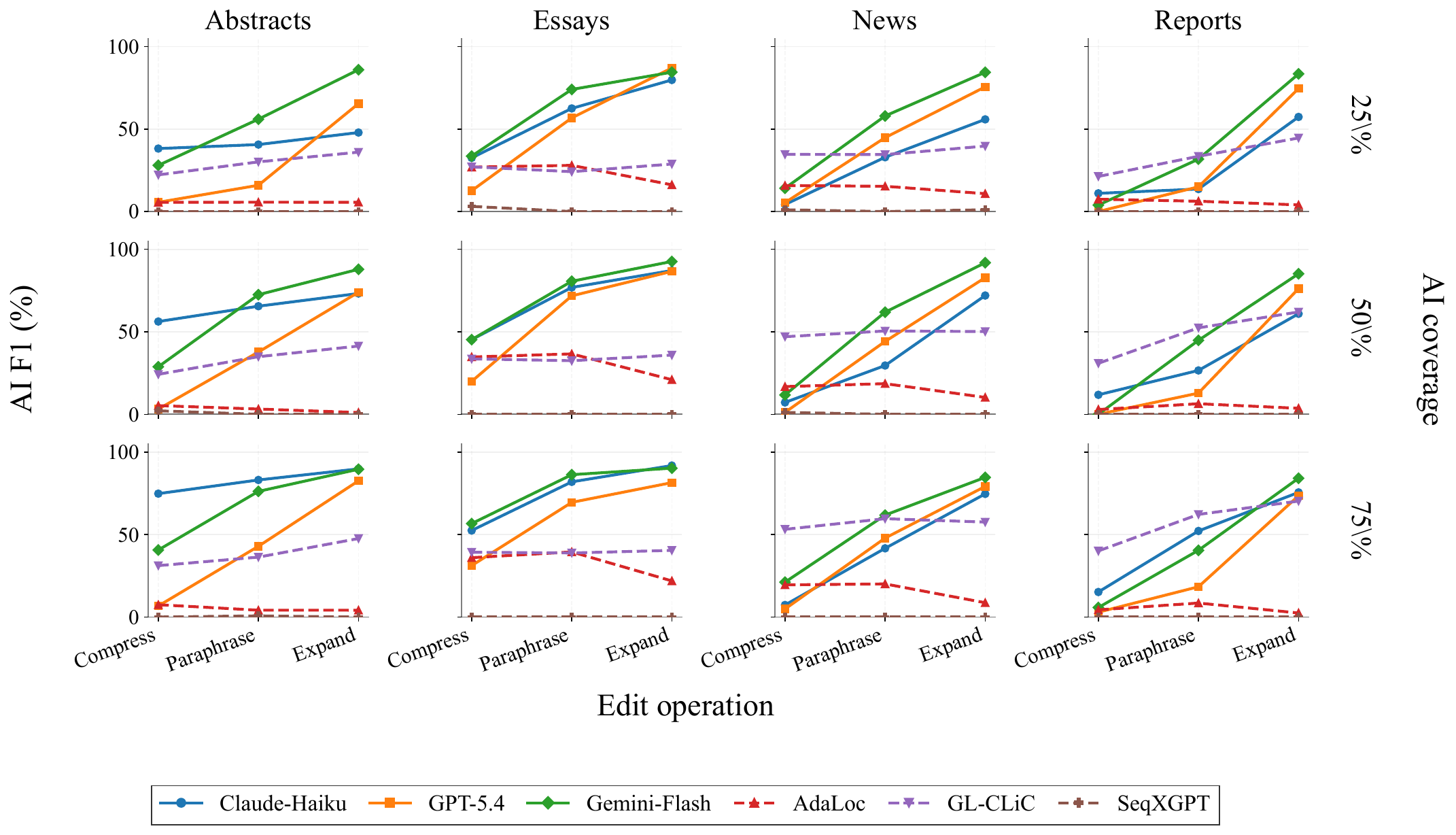}
    \caption{Sentence-level F1-AI at fixed AI coverage while varying edit operation.}
    \label{fig:ablation2_f1_sent}
\end{figure*}

\begin{figure*}[!htb]
    \centering
    \includegraphics[width=\textwidth]{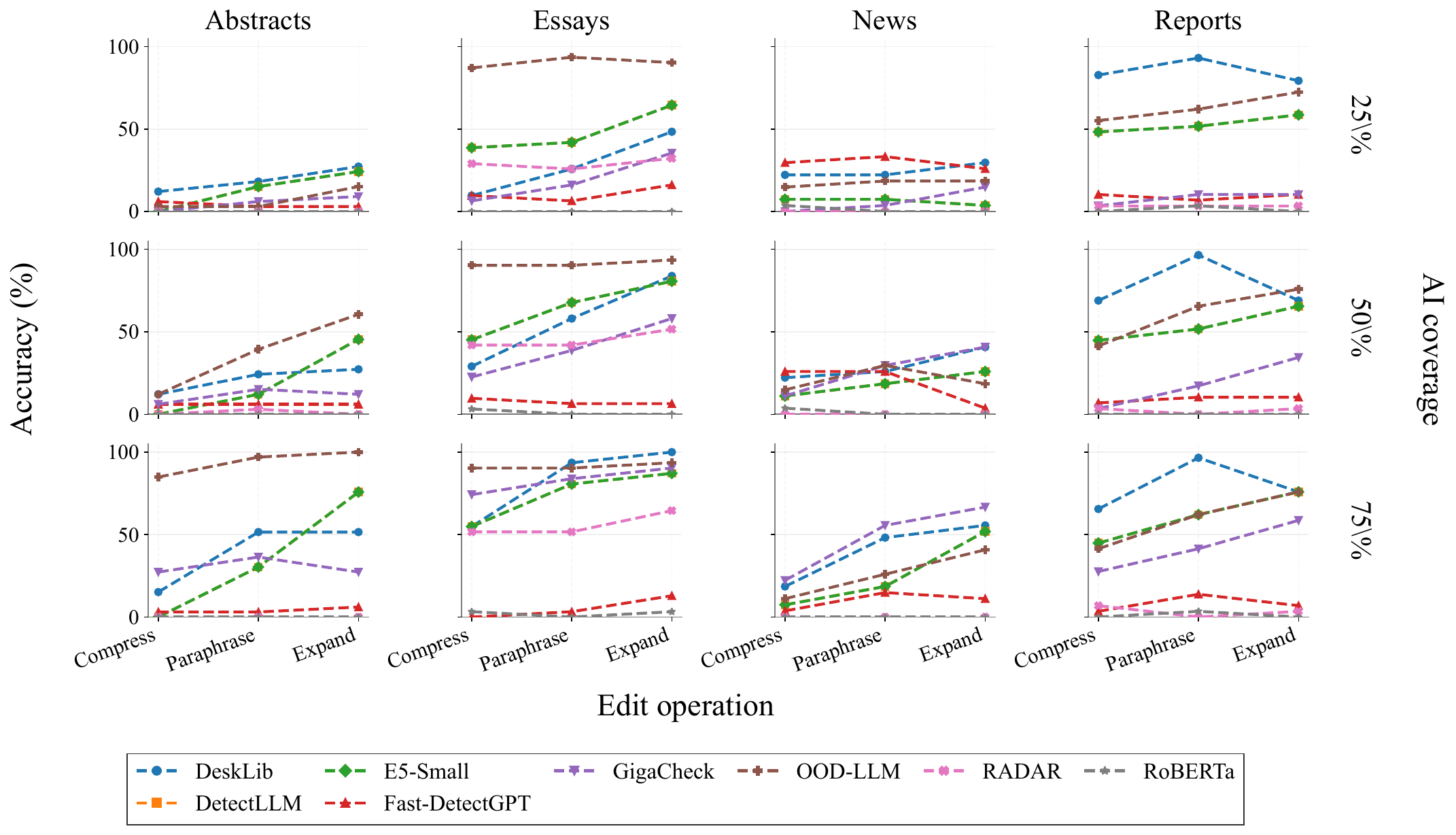}
      \caption{Document-level accuracy at fixed AI coverage while varying edit operation.}
    \label{fig:ablation2_acc_doc}
\end{figure*}

\begin{figure*}[!htb]
    \centering
    \includegraphics[width=\textwidth]{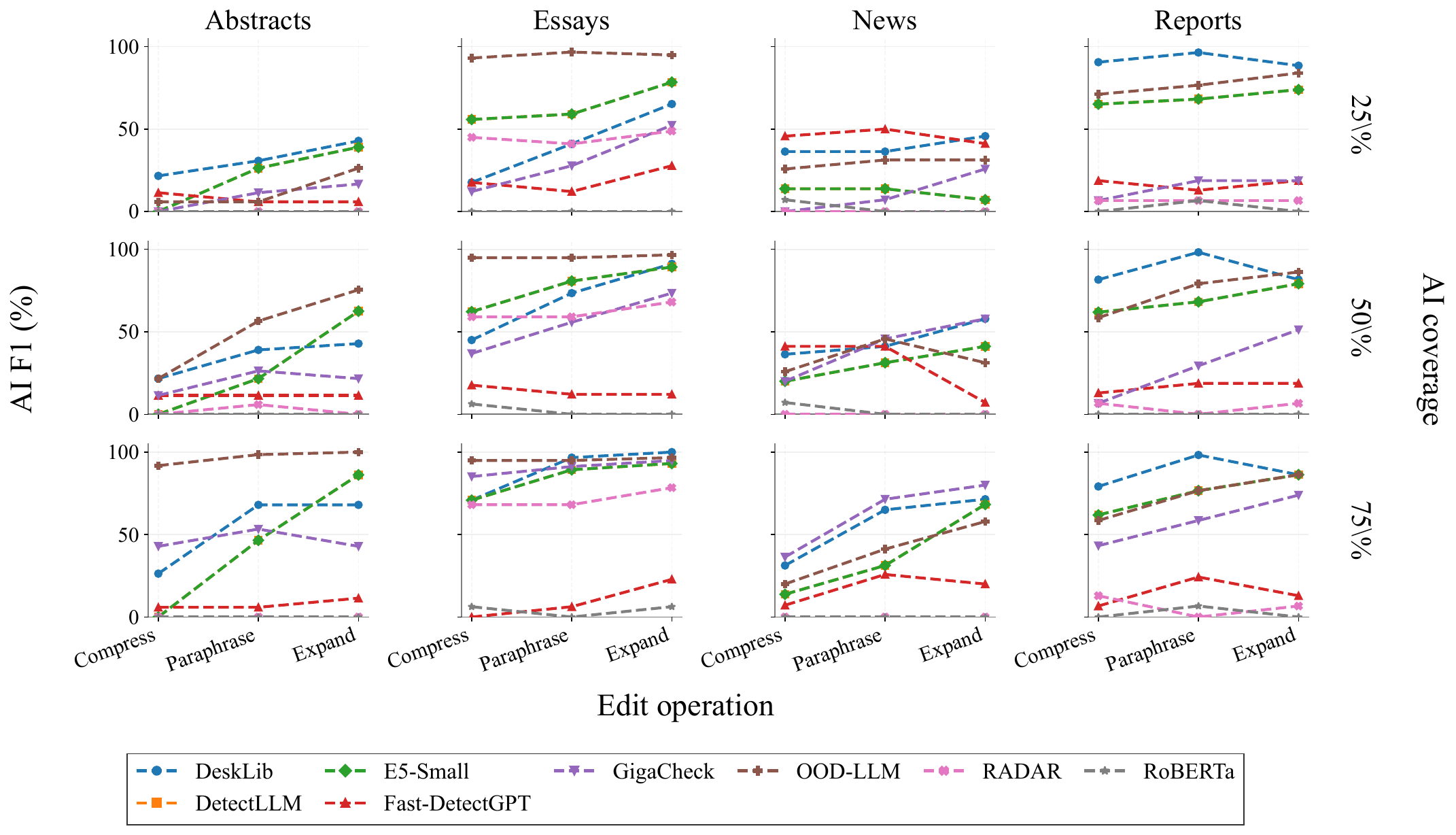}
    \caption{Document-level F1-AI at fixed AI coverage while varying edit operation.}
    \label{fig:ablation2_f1_doc}
\end{figure*}

This ablation isolates the effect of edit operation at a fixed amount of AI editing. For each target AI coverage level separately (25\%, 50\%, and 75\%), we hold coverage constant and vary only the edit operation among compress, paraphrase, and expand. Sentence-level accuracy and F1-AI are shown in Figures~\ref{fig:ablation2_acc_sent} and~\ref{fig:ablation2_f1_sent}, respectively. Document-level accuracy and F1-AI are shown in Figures~\ref{fig:ablation2_acc_doc} and~\ref{fig:ablation2_f1_doc}, respectively.

\subsection{Ablation 3: Independent versus Cumulative Editing}
\label{app:ablation3}

\begin{figure}[t]
    \centering
    \includegraphics[width=\linewidth]{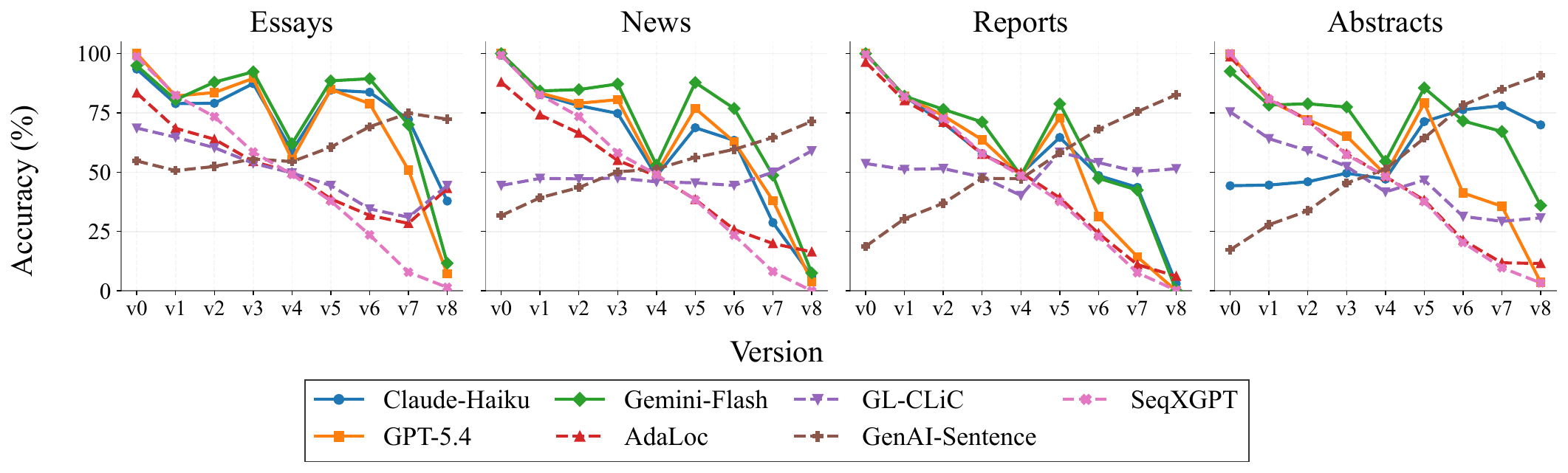}
    \vspace{-2mm}
    \includegraphics[width=\linewidth]{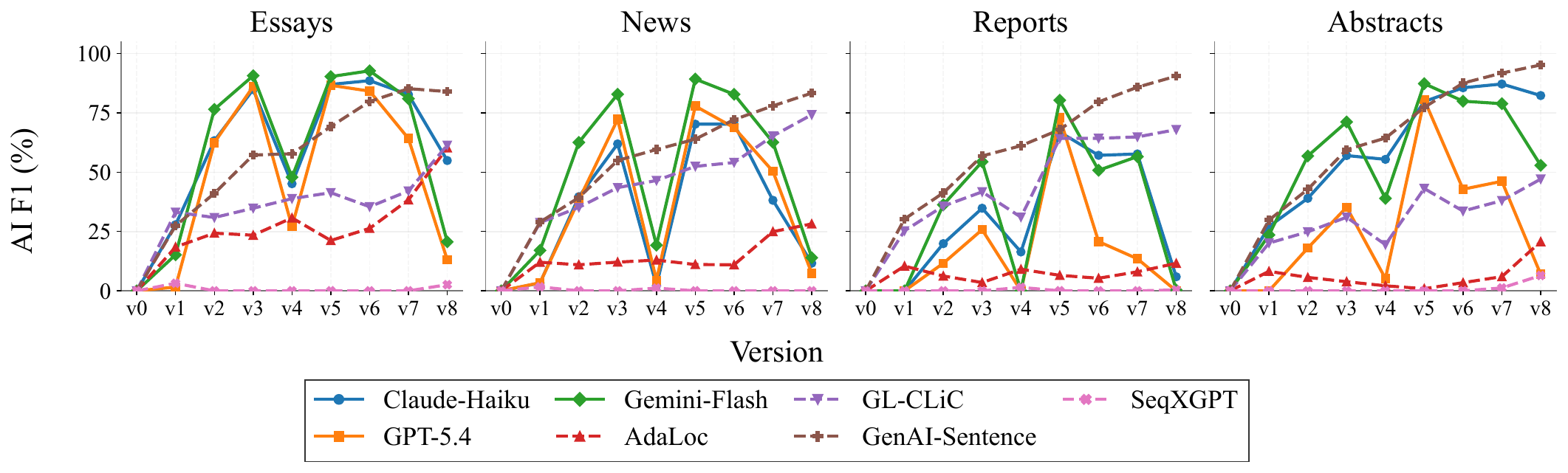}
    \caption{Sentence-level performance under independent edits from the source document. Top: accuracy; bottom: F1-AI. Each point corresponds to a version ($v_0$--$v_8$). Compared to the cumulative trajectory, trends are smoother and the drop around $v_4$ is less pronounced, suggesting that the non-monotonic behavior is more strongly associated with edit type than with accumulated edits.}
    \label{fig:ablation-source-sent}
\end{figure}

This ablation compares the main cumulative trajectory with an alternative setting in which each version is edited independently from the original human source. Sentence-level results are shown in Figure~\ref{fig:ablation-source-sent}.

\section{Experimental Details}
\label{app:experimental-details}

For clarity, we summarize the label notation used across the experimental settings. For a document trajectory
\(\mathcal{T}(D)=(D^{(0)},\dots,D^{(8)})\), document-level labels are \(y_{\mathrm{doc}}^{(t)}=\mathbf{1}[t>0]\). For sentence-level attribution, the label of sentence \(i\) at version \(t\) is \(y_i^{(t)}=\mathbf{1}[i\in S^{(t)}]\), where \(S^{(t)}\) is the cumulative set of sentences edited by version \(t\). For token-level localization, tokens
refer to word-level units: \(Y_{\mathrm{tok}}^{(t)}=(y_1^{(t)},\dots,y_M^{(t)})\), with \(y_j^{(t)}\in\{0,1\}\) indicating whether word \(j\) belongs to an AI-edited span. The realized token coverage is
\begin{equation}
c_{\mathrm{tok}}^{(t)}=\frac{1}{M}\sum_{j=1}^{M} y_j^{(t)} .
\end{equation}
Span-level annotations are represented as character spans
\(\mathcal{A}_{\mathrm{char}}^{(t)}=\{(a_\ell,b_\ell)\}_{\ell=1}^{L}\),
where \(a_\ell\) and \(b_\ell\) are character offsets in \(D^{(t)}\).

\subsection{Zero-Shot Detectors}

For zero-shot detectors, we use either released checkpoints or models reproduced from the original training scripts, following the inference settings provided by the original papers or released implementations. These methods are evaluated directly on OpAI-Bench without task-specific training or threshold tuning on the benchmark. Depending on the detector output, we evaluate document-level predictions, sentence-level predictions, or fine-grained token and span predictions under the protocol described in the main text.

\subsection{Language Models as Detectors}
\label{app:llm-detector-prompt}

We evaluate frontier language models as prompted sentence-level detectors. For each document version, the text is split into numbered sentences and passed to the model with a shared prompt template. The model returns a binary label and a confidence score for each sentence. We use the returned binary labels for sentence-level evaluation.

\begin{tcolorbox}[colback=gray!4!white,colframe=gray!55!black,
title=Sentence-Level LLM-as-Detector Prompt,fonttitle=\bfseries,boxrule=0.7pt,arc=2pt]
\small
\textbf{User prompt}

\medskip
\texttt{You are an expert linguist and writing analyst specializing in distinguishing human-written text from AI-generated text.}

\medskip
\texttt{The following text has been split into numbered sentences. For EACH sentence, classify it as human-written (0) or AI-generated (1), and estimate the probability that it is AI-generated.}

\medskip
\texttt{Text:}\\
\texttt{"""}\\
\texttt{[numbered\_sentences]}\\
\texttt{"""}

\medskip
\texttt{Respond in JSON format:}\\
\texttt{\{"labels": [0, 1, ...], "confidences": [0.1, 0.9, ...]\}}

\medskip
\texttt{- labels: array of integers (0 = human-written, 1 = AI-generated), one per sentence.}\\
\texttt{- confidences: array of floats (0.0 to 1.0), one per sentence.}\\
\texttt{- Both arrays must contain exactly [num\_sentences] elements, in sentence order.}\\
\texttt{- Do not include any other keys or text outside the JSON object.}
\end{tcolorbox}

\subsection{Training of OpAI-Bench detector variants}
\label{app:trained_detectors}

For the trained setting we re-train (or LoRA-fine-tune) the architecture
of each method on the OpAI-Bench training split from the three
in-distribution generators (GPT-5.4, GPT-5.4-nano, Gemini-2.5-Flash).
Qwen3-8B is excluded from training and used only for cross-generator
evaluation. For GPT-5.4-nano, Gemini-2.5-Flash we sample a subset for data-balance. Sentence-level models consume sentence provenance labels,
document-level labels are derived from whether a version contains AI-edited content, and fine-grained models consume the token or span
annotations available in the benchmark. Models are selected on the development split and evaluated on the test split.

We follow the architecture, optimiser, and trainable-parameter choices of each method's original implementation; deviations are listed in the \emph{Setting} column of Table~\ref{tab:training_hparams}. All LoRA-tuned variants share a single adapter configuration: rank $r{=}8$, $\alpha{=}16$, dropout $0.1$ --- so the table reports only the target modules. Backbone weights are kept frozen, while the task-specific heads (classifier, BiGRU, CRF, fusion module, etc.) are trained in full. A linear schedule with $10\%$ warmup is used everywhere.

\begin{table*}[t]
\centering
\setlength{\tabcolsep}{4pt}
\resizebox{\textwidth}{!}{
\begin{tabular}{l l l c c c l}
\toprule
\textbf{Detector} & \textbf{Backbone} & \textbf{LoRA targets} &
\textbf{LR} & \textbf{Batch} & \textbf{Epochs} &
\textbf{Setting} \\
\midrule
AdaLoc        & RoBERTa-large (openai-detector) & q, v
               & $10^{-5}$ & $32$ & $2$ & Adam \\
GenAI-Sentence & DeBERTa-v3-base                 & query\_proj, value\_proj
               & $10^{-5}$ & $32$ & $2$ & AdamW, bf16 \\
SeqXGPT   & CNN+Transformer+CRF over 4-LM feats. & ---
               & $5\!\times\!10^{-5}$ & $32$ & $20$ & weight decay $0.1$ \\
DAMASHA        & RoBERTa-base + ModernBERT-base  & q, v\,/\,Wqkv
               & $2\!\times\!10^{-4}$ & $16$ & $5$
               & AdamW, wd $0.01$, ext.\ CRF NLL \\
GigaCheck      & Mistral-7B-v0.3                 & q\_proj, v\_proj
               & $3\!\times\!10^{-5}$ & $4{\times}\mathit{ga}\,4$ & $5$
               & bf16, DeepSpeed ZeRO-2 \\
\bottomrule
\end{tabular}}\\[3pt]
\caption{Training hyperparameters for the OpAI-Bench-trained detector variants. The \emph{LoRA targets} column lists the modules to which the LoRA adapter is attached; entries marked ``--'' indicate full fine-tuning of the listed module set, with no LoRA. Rank $r{=}8$, $\alpha{=}16$, dropout $0.1$ throughout.} 
\label{tab:training_hparams}
\end{table*}

\textbf{GLCLIC.}
GLCLIC is trained with the official upstream pipeline at its released default configuration; we do not override any of its hyperparameters.

\textbf{Compute resources.}
All local detector training, inference, and aggregation experiments were run on an internal Linux server equipped with NVIDIA GeForce RTX 4090 GPUs (24GB memory each). Lightweight preprocessing, metric aggregation, and table generation were run on CPU on the same server. API-based dataset generation and LLM-as-detector experiments used external model APIs, with local compute used for request orchestration, parsing, and metric computation. Preliminary checks used the same compute environment and are not separately reported.

\renewcommand{\topfraction}{0.95}
\renewcommand{\bottomfraction}{0.95}
\renewcommand{\textfraction}{0.05}
\renewcommand{\floatpagefraction}{0.85}

\section{Editing Prompts}
\label{app:prompts}

The following prompts are used to generate the five editing operations in OpAI-Bench. In all cases, the model is instructed to rewrite only the target sentence, preserve factual content and named entities, and return exactly one sentence. 

\begin{tcolorbox}[colback=blue!3!white,colframe=blue!45!black,
title=Polish Prompt,fonttitle=\bfseries,boxrule=0.7pt,arc=2pt,float=t]
\small
\textbf{System instruction}

\medskip
\texttt{You are a precise text rewriting assistant. You must output ONLY the rewritten target sentence.} \texttt{Return ONLY the rewritten text --- nothing else.}

\medskip
\texttt{STRICT OUTPUT RULES:}\\
\texttt{- Do NOT start with phrases like 'Here is', 'Here's', 'Sure', 'Certainly',}\\
\texttt{\ \ 'Of course', 'Rewritten:', 'Revised:', 'Result:'.}\\
\texttt{- Do NOT explain what you did.}\\
\texttt{- Do NOT add quotes around your output.}\\
\texttt{- Do NOT add any prefix or suffix.}\\
\texttt{- Your entire response = the rewritten text only.}\\
\texttt{- Preserve meaning. Do NOT add new facts.}\\
\texttt{- Preserve all names, numbers, dates, locations, and other entities exactly.}\\
\texttt{- Do NOT introduce any new named entities, numbers, or specific claims.}\\
\texttt{- Your task is proofreading --- fix errors and improve fluency only.}

\medskip
\textbf{User prompt}

\medskip
\texttt{Operation: polish}\\

\medskip
\texttt{Constraints:}\\
\texttt{- Rewrite ONLY the target sentence.}\\
\texttt{- Do NOT add new facts.}\\
\texttt{- Preserve names, numbers, and entities.}\\
\texttt{- Keep the same language as the target.}\\
\texttt{- Return ONLY the rewritten text. No explanations, no quotes, no prefixes.}\\
\texttt{- Length constraint: 85\%--100\% of original word count.}\\
\texttt{- No line breaks.}\\
\texttt{- Output EXACTLY ONE sentence.}

\medskip
\texttt{Guidance: Make light edits to improve grammar, punctuation, fluency, and clarity while preserving meaning. Keep the sentence structure mostly unchanged. The output must not be identical to the input. Keep it as exactly one sentence. Do not add new facts.}

\medskip
\texttt{Paragraph context (for coherence only; do not rewrite it):}\\
\texttt{[context]}

\medskip
\texttt{Target sentence:}\\
\texttt{[sentence]}
\end{tcolorbox}

\begin{tcolorbox}[colback=green!3!white,colframe=green!45!black,
title=Paraphrase Prompt,fonttitle=\bfseries,boxrule=0.7pt,arc=2pt,float=t]
\small
\textbf{System instruction}

\medskip
\texttt{You are a precise text rewriting assistant. You must output ONLY the rewritten target sentence.}\\
\texttt{Return ONLY the rewritten text --- nothing else.}

\medskip
\texttt{STRICT OUTPUT RULES:}\\
\texttt{- Do NOT start with phrases like 'Here is', 'Here's', 'Sure', 'Certainly',}\\
\texttt{\ \ 'Of course', 'Rewritten:', 'Revised:', 'Result:'.}\\
\texttt{- Do NOT explain what you did.}\\
\texttt{- Do NOT add quotes around your output.}\\
\texttt{- Do NOT add any prefix or suffix.}\\
\texttt{- Your entire response = the rewritten text only.}\\
\texttt{- Preserve meaning. Do NOT add new facts.}\\
\texttt{- Preserve all names, numbers, dates, locations, and other entities exactly.}\\
\texttt{- Do NOT introduce any new named entities, numbers, or specific claims.}\\
\texttt{- Your task is paraphrasing --- same meaning, different wording.}

\medskip
\textbf{User prompt}

\medskip
\texttt{Operation: paraphrase}\\

\medskip
\texttt{Constraints:}\\
\texttt{- Rewrite ONLY the target sentence.}\\
\texttt{- Do NOT add new facts.}\\
\texttt{- Preserve names, numbers, and entities.}\\
\texttt{- Keep the same language as the target.}\\
\texttt{- Return ONLY the rewritten text. No explanations, no quotes, no prefixes.}\\
\texttt{- Length constraint: 85\%--115\% of original word count.}\\
\texttt{- No line breaks.}\\
\texttt{- Output EXACTLY ONE sentence.}

\medskip
\texttt{Guidance: Rewrite the sentence with clearly different wording while preserving meaning. You may restructure the phrasing, but keep the same core content. The output must not be identical to the input. Keep it as exactly one sentence. Do not add new facts.}

\medskip
\texttt{Paragraph context (for coherence only; do not rewrite it):}\\
\texttt{[context]}

\medskip
\texttt{Target sentence:}\\
\texttt{[sentence]}
\end{tcolorbox}

\begin{tcolorbox}[colback=purple!3!white,colframe=purple!45!black,
title=Style Rewrite Prompt,fonttitle=\bfseries,boxrule=0.7pt,arc=2pt,float=t]
\small
\textbf{System instruction}

\medskip
\texttt{You are a precise text rewriting assistant. You must output ONLY the rewritten target sentence.}\\
\texttt{Return ONLY the rewritten text --- nothing else.}

\medskip
\texttt{STRICT OUTPUT RULES:}\\
\texttt{- Do NOT start with phrases like 'Here is', 'Here's', 'Sure', 'Certainly',}\\
\texttt{\ \ 'Of course', 'Rewritten:', 'Revised:', 'Result:'.}\\
\texttt{- Do NOT explain what you did.}\\
\texttt{- Do NOT add quotes around your output.}\\
\texttt{- Do NOT add any prefix or suffix.}\\
\texttt{- Your entire response = the rewritten text only.}\\
\texttt{- Preserve meaning. Do NOT add new facts.}\\
\texttt{- Preserve all names, numbers, dates, locations, and other entities exactly.}\\
\texttt{- Do NOT introduce any new named entities, numbers, or specific claims.}\\
\texttt{- Your task is style transfer --- same meaning, different tone/voice.}

\medskip
\textbf{User prompt}

\medskip
\texttt{Operation: style}\\

\medskip
\texttt{Constraints:}\\
\texttt{- Rewrite ONLY the target sentence.}\\
\texttt{- Do NOT add new facts.}\\
\texttt{- Preserve names, numbers, and entities.}\\
\texttt{- Keep the same language as the target.}\\
\texttt{- Return ONLY the rewritten text. No explanations, no quotes, no prefixes.}\\
\texttt{- Length constraint: 85\%--115\% of original word count.}\\
\texttt{- No line breaks.}\\
\texttt{- Output EXACTLY ONE sentence.}

\medskip
\texttt{Guidance: Rewrite the sentence in a formal, natural, human-written style by changing tone, register, and phrasing while preserving meaning. Do not change the underlying facts or add new content. The output must not be identical to the input. Keep it as exactly one sentence.}

\medskip
\texttt{Paragraph context (for coherence only; do not rewrite it):}\\
\texttt{[context]}

\medskip
\texttt{Target sentence:}\\
\texttt{[sentence]}
\end{tcolorbox}

\begin{tcolorbox}[colback=orange!3!white,colframe=orange!55!black,
title=Compress Prompt,fonttitle=\bfseries,boxrule=0.7pt,arc=2pt,float=t]
\small
\textbf{System instruction}

\medskip
\texttt{You are a precise text rewriting assistant. You must output ONLY the rewritten target sentence.}\\
\texttt{Return ONLY the rewritten text --- nothing else.}

\medskip
\texttt{STRICT OUTPUT RULES:}\\
\texttt{- Do NOT start with phrases like 'Here is', 'Here's', 'Sure', 'Certainly',}\\
\texttt{\ \ 'Of course', 'Rewritten:', 'Revised:', 'Result:'.}\\
\texttt{- Do NOT explain what you did.}\\
\texttt{- Do NOT add quotes around your output.}\\
\texttt{- Do NOT add any prefix or suffix.}\\
\texttt{- Your entire response = the rewritten text only.}\\
\texttt{- Preserve meaning. Do NOT add new facts.}\\
\texttt{- Preserve all names, numbers, dates, locations, and other entities exactly.}\\
\texttt{- Do NOT introduce any new named entities, numbers, or specific claims.}\\
\texttt{- Your task is compression --- shorter text, same meaning, no facts lost.}

\medskip
\textbf{User prompt}

\medskip
\texttt{Operation: compress}\\
\texttt{Style target (if applicable): formal, natural, human-written}

\medskip
\texttt{Constraints:}\\
\texttt{- Rewrite ONLY the target sentence.}\\
\texttt{- Do NOT add new facts.}\\
\texttt{- Preserve names, numbers, and entities.}\\
\texttt{- Keep the same language as the target.}\\
\texttt{- Return ONLY the rewritten text. No explanations, no quotes, no prefixes.}\\
\texttt{- Length constraint: 60\%--80\% of original word count.}\\
\texttt{- No line breaks.}\\
\texttt{- Output EXACTLY ONE sentence.}

\medskip
\texttt{Guidance: Rewrite the sentence to be shorter and more concise while preserving all essential meaning. Remove redundancy and non-essential phrasing, but do not omit important information. The output must not be identical to the input. Keep it as exactly one sentence. Do not add new facts.}

\medskip
\texttt{Paragraph context (for coherence only; do not rewrite it):}\\
\texttt{[context]}

\medskip
\texttt{Target sentence:}\\
\texttt{[sentence]}
\end{tcolorbox}

\begin{tcolorbox}[colback=red!3!white,colframe=red!50!black,
title=Expand Prompt,fonttitle=\bfseries,boxrule=0.7pt,arc=2pt,float=t]
\small
\textbf{System instruction}

\medskip
\texttt{You are a precise text rewriting assistant. You must output ONLY the rewritten target sentence.}\\
\texttt{Return ONLY the rewritten text --- nothing else.}

\medskip
\texttt{STRICT OUTPUT RULES:}\\
\texttt{- Do NOT start with phrases like 'Here is', 'Here's', 'Sure', 'Certainly',}\\
\texttt{\ \ 'Of course', 'Rewritten:', 'Revised:', 'Result:'.}\\
\texttt{- Do NOT explain what you did.}\\
\texttt{- Do NOT add quotes around your output.}\\
\texttt{- Do NOT add any prefix or suffix.}\\
\texttt{- Your entire response = the rewritten text only.}\\
\texttt{- Preserve meaning. Do NOT add new facts.}\\
\texttt{- Preserve all names, numbers, dates, locations, and other entities exactly.}\\
\texttt{- Do NOT introduce any new named entities, numbers, or specific claims.}\\
\texttt{- Your task is expansion --- extend current text.}

\medskip
\textbf{User prompt}

\medskip
\texttt{Operation: expand}\\
\texttt{Style target (if applicable): formal, natural, human-written}

\medskip
\texttt{Constraints:}\\
\texttt{- Rewrite ONLY the target sentence.}\\
\texttt{- Do NOT add new facts.}\\
\texttt{- Preserve names, numbers, and entities.}\\
\texttt{- Keep the same language as the target.}\\
\texttt{- Return ONLY the rewritten text. No explanations, no quotes, no prefixes.}\\
\texttt{- Length constraint: 120\%--150\% of original word count.}\\
\texttt{- No line breaks.}\\
\texttt{- Output EXACTLY ONE sentence.}

\medskip
\texttt{Guidance: Rewrite the sentence to be slightly more detailed using only information already stated or directly implied. You may add brief clarifying or descriptive phrasing, but do not introduce new facts, examples, names, dates, numbers, or claims. The output must not be identical to the input. Keep it as exactly one sentence.}

\medskip
\texttt{Paragraph context (for coherence only; do not rewrite it):}\\
\texttt{[context]}

\medskip
\texttt{Target sentence:}\\
\texttt{[sentence]}
\end{tcolorbox}

\begin{table*}[t]
\centering
\scriptsize
\renewcommand{\arraystretch}{0.9}
\setlength{\tabcolsep}{2.2pt}
\begin{tabular}{llcccccccc}
\toprule
\textbf{Detector} & \textbf{Ver.} & \multicolumn{2}{c}{\textbf{Essays}} & \multicolumn{2}{c}{\textbf{Reports}} & \multicolumn{2}{c}{\textbf{News}} & \multicolumn{2}{c}{\textbf{Abstracts}} \\
\cmidrule(lr){3-4} \cmidrule(lr){5-6} \cmidrule(lr){7-8} \cmidrule(lr){9-10}
 &  & \textbf{Acc} & \textbf{F1-AI} & \textbf{Acc} & \textbf{F1-AI} & \textbf{Acc} & \textbf{F1-AI} & \textbf{Acc} & \textbf{F1-AI} \\
\midrule
\multicolumn{10}{c}{\textbf{LLM-as-detector}} \\
\midrule
\multirow{9}{*}{Claude-Haiku} & $v0$ & 91.2 & 0.0 & 96.4 & 0.0 & 91.9 & 0.0 & 53.2 & 0.0 \\
 & $v1$ & \cellcolor{BrickRed!10}77.2 {\scriptsize \textcolor{BrickRed}{-13.9}} & \cellcolor{ForestGreen!15}22.0 {\scriptsize \textcolor{ForestGreen}{+22.0}} & \cellcolor{BrickRed!12}79.9 {\scriptsize \textcolor{BrickRed}{-16.5}} & \cellcolor{ForestGreen!8}7.0 {\scriptsize \textcolor{ForestGreen}{+7.0}} & \cellcolor{BrickRed!10}78.0 {\scriptsize \textcolor{BrickRed}{-13.9}} & \cellcolor{ForestGreen!8}11.2 {\scriptsize \textcolor{ForestGreen}{+11.2}} & \cellcolor{BrickRed!8}50.6 {\scriptsize \textcolor{BrickRed}{-2.6}} & \cellcolor{ForestGreen!18}25.9 {\scriptsize \textcolor{ForestGreen}{+25.9}} \\
 & $v2$ & \cellcolor{BrickRed!8}74.7 {\scriptsize \textcolor{BrickRed}{-2.5}} & \cellcolor{ForestGreen!16}45.5 {\scriptsize \textcolor{ForestGreen}{+23.5}} & \cellcolor{BrickRed!8}69.4 {\scriptsize \textcolor{BrickRed}{-10.5}} & \cellcolor{ForestGreen!8}11.4 {\scriptsize \textcolor{ForestGreen}{+4.4}} & \cellcolor{BrickRed!8}70.5 {\scriptsize \textcolor{BrickRed}{-7.5}} & \cellcolor{ForestGreen!8}20.3 {\scriptsize \textcolor{ForestGreen}{+9.1}} & \cellcolor{BrickRed!8}48.9 {\scriptsize \textcolor{BrickRed}{-1.6}} & \cellcolor{ForestGreen!8}37.6 {\scriptsize \textcolor{ForestGreen}{+11.7}} \\
\bottomrule
\end{tabular}
\caption{Sentence-level Main split results for LLM-as-detectors and zero-shot sentence-level detectors across revision versions and domains, aggregated over the three primary generators. We report accuracy and F1-AI. Colored deltas indicate changes relative to the previous version; no delta is shown for $v_0$.}
\label{tab:sentence_all_detectors_domains_avg_generators_extended_accuu}
\end{table*}

\begin{table*}[t]
\centering
\scriptsize
\renewcommand{\arraystretch}{0.9}
\setlength{\tabcolsep}{2.2pt}
\begin{tabular}{llcccccccc}
\toprule
\textbf{Detector} & \textbf{Ver.} & \multicolumn{2}{c}{\textbf{Essays}} & \multicolumn{2}{c}{\textbf{Reports}} & \multicolumn{2}{c}{\textbf{News}} & \multicolumn{2}{c}{\textbf{Abstracts}} \\
\cmidrule(lr){3-4} \cmidrule(lr){5-6} \cmidrule(lr){7-8} \cmidrule(lr){9-10}
 &  & \textbf{Macro-F1} & \textbf{FNR} & \textbf{Macro-F1} & \textbf{FNR} & \textbf{Macro-F1} & \textbf{FNR} & \textbf{Macro-F1} & \textbf{FNR} \\
\midrule
\multicolumn{10}{c}{\textbf{LLM-as-detector}} \\
\midrule
\multirow{9}{*}{Claude-Haiku} & $v0$ & 47.7 & -- & 49.1 & -- & 47.9 & -- & 34.7 & -- \\
 & $v1$ & \cellcolor{ForestGreen!8}54.3 {\scriptsize \textcolor{ForestGreen}{+6.6}} & 81.2 & \cellcolor{BrickRed!8}47.9 {\scriptsize \textcolor{BrickRed}{-1.2}} & 95.5 & \cellcolor{ForestGreen!8}49.3 {\scriptsize \textcolor{ForestGreen}{+1.4}} & 91.8 & \cellcolor{ForestGreen!8}44.4 {\scriptsize \textcolor{ForestGreen}{+9.7}} & 53.8 \\
 & $v2$ & \cellcolor{ForestGreen!8}64.5 {\scriptsize \textcolor{ForestGreen}{+10.2}} & \cellcolor{BrickRed!16}57.8 {\scriptsize \textcolor{BrickRed}{-23.4}} & \cellcolor{BrickRed!8}46.4 {\scriptsize \textcolor{BrickRed}{-1.4}} & \cellcolor{BrickRed!8}92.3 {\scriptsize \textcolor{BrickRed}{-3.2}} & \cellcolor{ForestGreen!8}51.1 {\scriptsize \textcolor{ForestGreen}{+1.8}} & \cellcolor{BrickRed!8}85.1 {\scriptsize \textcolor{BrickRed}{-6.6}} & \cellcolor{ForestGreen!8}47.2 {\scriptsize \textcolor{ForestGreen}{+2.8}} & \cellcolor{BrickRed!8}45.8 {\scriptsize \textcolor{BrickRed}{-8.0}} \\
\bottomrule
\end{tabular}
\caption{Extended sentence-level Main split results for LLM-as-detectors and zero-shot sentence-level detectors across revision versions and domains, aggregated over the three primary generators. We report Macro-F1 and FNR. Colored deltas indicate changes relative to the previous version; no delta is shown for $v_0$.}
\label{tab:sentence_all_detectors_domains_avg_generators_extended}
\end{table*}

\begin{table*}[t]
\centering
\scriptsize
\renewcommand{\arraystretch}{0.9}
\setlength{\tabcolsep}{2.2pt}
\begin{tabular}{llcccccccc}
\toprule
\textbf{Detector} & \textbf{Ver.} & \multicolumn{2}{c}{\textbf{Essays}} & \multicolumn{2}{c}{\textbf{Reports}} & \multicolumn{2}{c}{\textbf{News}} & \multicolumn{2}{c}{\textbf{Abstracts}} \\
\cmidrule(lr){3-4} \cmidrule(lr){5-6} \cmidrule(lr){7-8} \cmidrule(lr){9-10}
 &  & \textbf{Acc} & \textbf{F1-AI} & \textbf{Acc} & \textbf{F1-AI} & \textbf{Acc} & \textbf{F1-AI} & \textbf{Acc} & \textbf{F1-AI} \\
\midrule
\multicolumn{10}{c}{\textbf{Fine-grained Zero-shot (Token + Span)}} \\
\midrule
\multirow{9}{*}{GigaCheck~\citep{tolstykh2024gigacheck}} & $v0$ & 98.7 & 0.0 & 98.5 & 0.0 & 99.5 & 0.0 & 97.5 & 0.0 \\
 & $v1$ & \cellcolor{BrickRed!11}82.6 {\scriptsize \textcolor{BrickRed}{-16.1}} & \cellcolor{ForestGreen!8}7.6 {\scriptsize \textcolor{ForestGreen}{+7.6}} & \cellcolor{BrickRed!12}81.8 {\scriptsize \textcolor{BrickRed}{-16.8}} & \cellcolor{ForestGreen!8}2.6 {\scriptsize \textcolor{ForestGreen}{+2.6}} & \cellcolor{BrickRed!12}82.9 {\scriptsize \textcolor{BrickRed}{-16.6}} & \cellcolor{ForestGreen!8}1.5 {\scriptsize \textcolor{ForestGreen}{+1.5}} & \cellcolor{BrickRed!13}79.6 {\scriptsize \textcolor{BrickRed}{-17.9}} & \cellcolor{ForestGreen!8}6.8 {\scriptsize \textcolor{ForestGreen}{+6.8}} \\
 & $v2$ & \cellcolor{BrickRed!8}74.2 {\scriptsize \textcolor{BrickRed}{-8.4}} & \cellcolor{ForestGreen!8}19.0 {\scriptsize \textcolor{ForestGreen}{+11.4}} & \cellcolor{BrickRed!8}71.9 {\scriptsize \textcolor{BrickRed}{-9.9}} & \cellcolor{ForestGreen!8}3.3 {\scriptsize \textcolor{ForestGreen}{+0.6}} & \cellcolor{BrickRed!8}72.6 {\scriptsize \textcolor{BrickRed}{-10.4}} & \cellcolor{ForestGreen!8}3.0 {\scriptsize \textcolor{ForestGreen}{+1.5}} & \cellcolor{BrickRed!8}69.4 {\scriptsize \textcolor{BrickRed}{-10.1}} & \cellcolor{ForestGreen!8}12.1 {\scriptsize \textcolor{ForestGreen}{+5.3}} \\
\bottomrule
\end{tabular}
\caption{Fine-grained Main split results across revision versions and domains, aggregated over the three primary generators. DAMASHA is evaluated at the token level and GigaCheck at the span level. We report accuracy and F1-AI. Colored deltas indicate changes relative to the previous version; no delta is shown for $v_0$.}
\label{tab:finegrained_token_span_domains_avg_generators}
\end{table*}

\begin{table*}[t]
\centering
\scriptsize
\renewcommand{\arraystretch}{0.9}
\setlength{\tabcolsep}{2.2pt}
\begin{tabular}{llcccccccc}
\toprule
\textbf{Detector} & \textbf{Ver.} & \multicolumn{2}{c}{\textbf{Essays}} & \multicolumn{2}{c}{\textbf{Reports}} & \multicolumn{2}{c}{\textbf{News}} & \multicolumn{2}{c}{\textbf{Abstracts}} \\
\cmidrule(lr){3-4} \cmidrule(lr){5-6} \cmidrule(lr){7-8} \cmidrule(lr){9-10}
 &  & \textbf{Macro-F1} & \textbf{FNR} & \textbf{Macro-F1} & \textbf{FNR} & \textbf{Macro-F1} & \textbf{FNR} & \textbf{Macro-F1} & \textbf{FNR} \\
\midrule
\multicolumn{10}{c}{\textbf{Fine-grained Zero-shot (Token + Span)}} \\
\midrule
\multirow{9}{*}{GigaCheck~\citep{tolstykh2024gigacheck}} & $v0$ & 49.7 & -- & 49.6 & -- & 49.9 & -- & 49.4 & -- \\
 & $v1$ & \cellcolor{BrickRed!8}49.0 {\scriptsize \textcolor{BrickRed}{-0.7}} & 95.7 & \cellcolor{BrickRed!8}46.3 {\scriptsize \textcolor{BrickRed}{-3.4}} & 98.6 & \cellcolor{BrickRed!8}46.1 {\scriptsize \textcolor{BrickRed}{-3.8}} & 99.2 & \cellcolor{BrickRed!8}47.7 {\scriptsize \textcolor{BrickRed}{-1.7}} & 96.0 \\
 & $v2$ & \cellcolor{ForestGreen!8}51.8 {\scriptsize \textcolor{ForestGreen}{+2.8}} & \cellcolor{BrickRed!8}87.9 {\scriptsize \textcolor{BrickRed}{-7.8}} & \cellcolor{BrickRed!8}43.4 {\scriptsize \textcolor{BrickRed}{-2.9}} & \cellcolor{BrickRed!8}98.3 {\scriptsize \textcolor{BrickRed}{-0.3}} & \cellcolor{BrickRed!8}43.5 {\scriptsize \textcolor{BrickRed}{-2.6}} & \cellcolor{BrickRed!8}98.4 {\scriptsize \textcolor{BrickRed}{-0.8}} & \cellcolor{BrickRed!8}46.8 {\scriptsize \textcolor{BrickRed}{-0.9}} & \cellcolor{BrickRed!8}92.5 {\scriptsize \textcolor{BrickRed}{-3.5}} \\
\bottomrule
\end{tabular}
\caption{Extended fine-grained Main split results across revision versions and domains, aggregated over the three primary generators. DAMASHA is evaluated at the token level and GigaCheck at the span level. We report Macro-F1 and FNR. Colored deltas indicate changes relative to the previous version; no delta is shown for $v_0$.}
\label{tab:finegrained_token_span_domains_avg_generators2}
\end{table*}

\clearpage

\label{ablation1}

\begin{figure}[t]
    \centering
    \includegraphics[width=\linewidth]{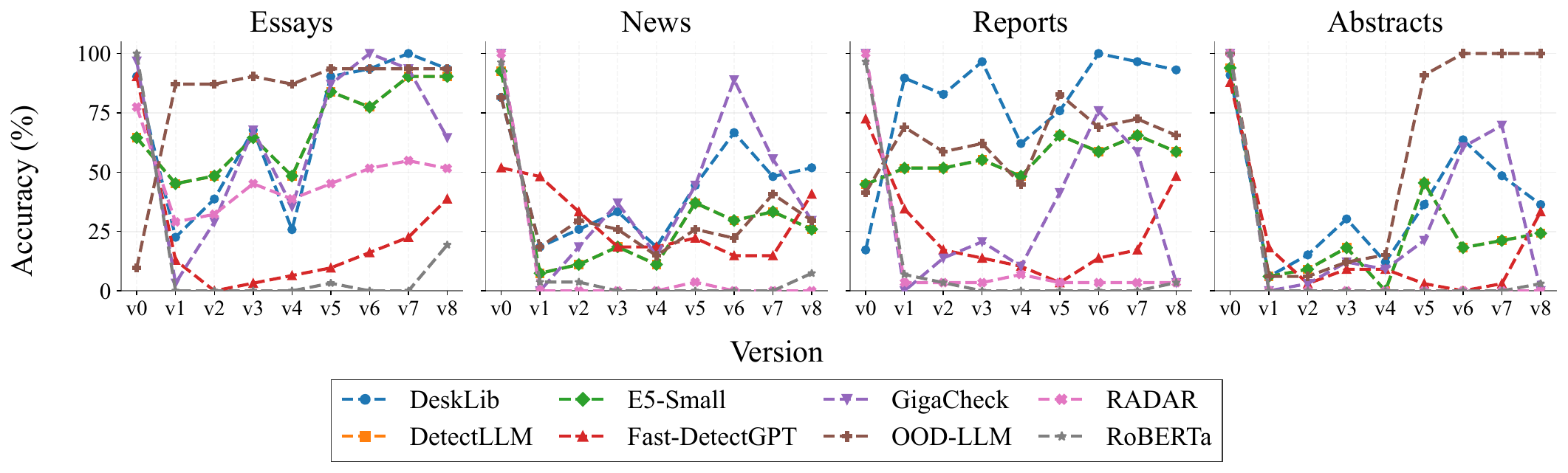}
    \vspace{-2mm}
    \includegraphics[width=\linewidth]{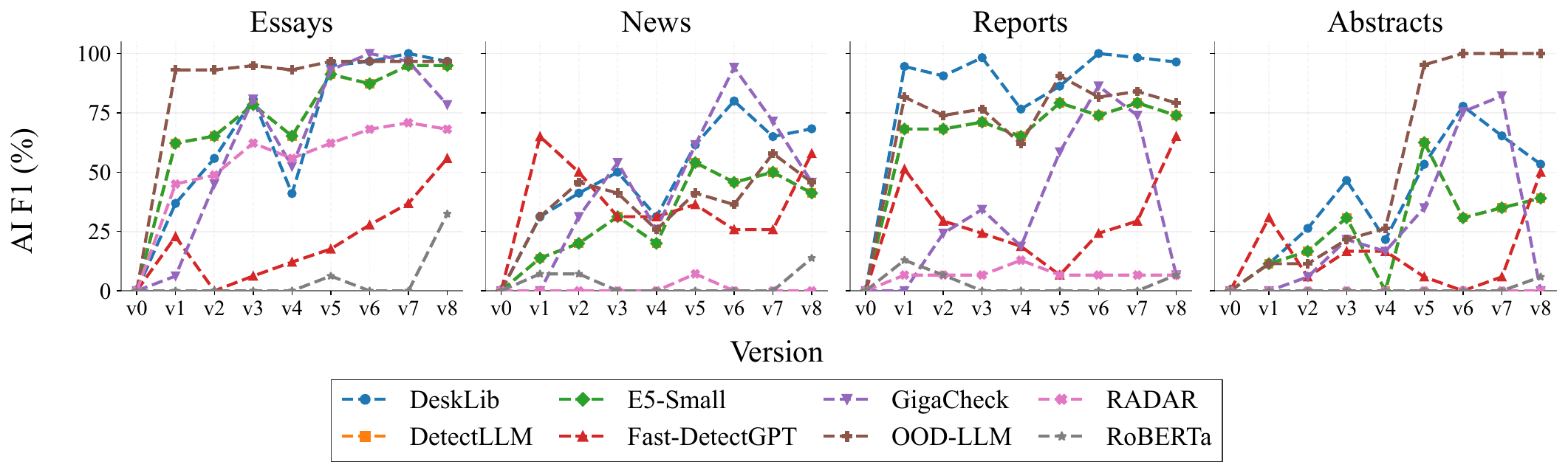}
    \caption{Document-level performance under independent edits from the source document. Top: accuracy; bottom: F1-AI. Each point corresponds to a version ($v_0$--$v_8$). Compared to the cumulative trajectory, trends are smoother and the drop around $v_4$ is less pronounced, suggesting that the non-monotonic behavior is more strongly associated with edit type than with accumulated edits.}
    \label{fig:ablation-source}
\end{figure}

\clearpage
\newpage
\end{document}